\documentclass[10pt,twocolumn,letterpaper]{article}

\usepackage[pagenumbers]{cvpr} 
\usepackage{multirow}
\usepackage{textcomp}
\usepackage{graphicx}
\usepackage{stfloats}
\usepackage{array}
\usepackage{amsmath,amsfonts}

\definecolor{cvprblue}{rgb}{0.21,0.49,0.74}
\usepackage[pagebackref,breaklinks,colorlinks,allcolors=cvprblue]{hyperref}

\def\paperID{15813} 
\def\confName{CVPR}
\def\confYear{2025}

\title{MaIR: A Locality- and Continuity-Preserving Mamba for Image Restoration}

\author{%
    Boyun Li\textsuperscript{1}, ~~%
    Haiyu Zhao\textsuperscript{1}, ~~%
    Wenxin Wang\textsuperscript{1}, ~~%
    Peng Hu\textsuperscript{1}, ~~%
    Yuanbiao Gou\textsuperscript{1}$^*$, ~~%
    Xi Peng\textsuperscript{1,2}\thanks{Corresponding authors} ~~%
    \\
    \textsuperscript{1} College of Computer Science, Sichuan University, China.\\
 \textsuperscript{2} National Key Laboratory of Fundamental Algorithms and Models for Engineering \\ Numerical Simulation, Sichuan University, China. \\
{\tt\small \{liboyun.gm, haiyuzhao.gm, wangwenxin.gm, penghu.ml, gouyuanbiao, pengx.gm\}@gmail.com}
}

\begin{document}
\maketitle

\begin{abstract}
Recent advancements in Mamba have shown promising results in image restoration. These methods typically flatten 2D images into multiple distinct 1D sequences along rows and columns, process each sequence independently using selective scan operation, and recombine them to form the outputs. However, such a paradigm overlooks two vital aspects: i) the local relationships and spatial continuity inherent in natural images, and ii) the discrepancies among sequences unfolded through totally different ways. To overcome the drawbacks, we explore two problems in Mamba-based restoration methods: i) how to design a scanning strategy preserving both locality and continuity while facilitating restoration, and ii) how to aggregate the distinct sequences unfolded in totally different ways. To address these problems, we propose a novel \underline{Ma}mba-based \underline{I}mage \underline{R}estoration model (MaIR), which consists of Nested S-shaped Scanning strategy (NSS) and Sequence Shuffle Attention block (SSA). Specifically, NSS preserves locality and continuity of the input images through the stripe-based scanning region and the S-shaped scanning path, respectively. SSA aggregates sequences through calculating attention weights within the corresponding channels of different sequences. Thanks to NSS and SSA, MaIR surpasses 40 baselines across 14 challenging datasets, achieving state-of-the-art performance on the tasks of image super-resolution, denoising, deblurring and dehazing. The code is available at \url{https://github.com/XLearning-SCU/2025-CVPR-MaIR}.
\end{abstract}

\begin{figure}[t!]
\vspace{0.5em}  
\begin{center}
\includegraphics[width=0.85\columnwidth]{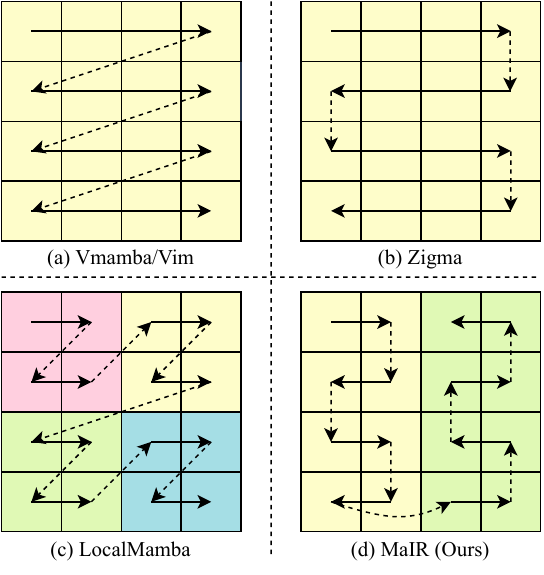}
\end{center}
\vspace{-2em}   
\caption{The scanning strategies in existing Mamba-based methods and our proposed method. (a) \textbf{Vmamba/Vim} uses Z-shaped scan path to flatten 2D image into 1D sequences, in which both the locality and continuity of 2D image are disrupted. (b) \textbf{Zigma} utilizes S-shaped path to maintain spatial continuity, while ignores the locality. (c) \textbf{LocalMamba} leverages window-based scanning region to preserve locality. However, the Z-shaped scanning path within and across the windows disrupts the spatial continuity. In contrast, (d) \textbf{MaIR} divides images into multiple non-overlapping stripes, and adopts S-shaped scanning path within and across the stripes, thus simultaneously preserves both locality and continuity.}
\vspace{-1em}
\label{Figure:observation}
\end{figure}

\section{Introduction}
Image restoration aims to recover visually appealing high-quality images from given degraded correspondences, \textit{e.g.}, noisy, blurry, and hazy images. In recent years, the methods based on Convolutional Neural Networks (CNNs) and Transformers have significantly advanced image restoration by effectively capturing locality (\textit{i.e.}, fine-grained patterns and correlations in small regions) and continuity (\textit{i.e.}, smooth, gradual transitions across larger areas) inherent in 2D natural images. To be specific, CNNs capture locality and continuity through the elaborately designed small kernels and sliding strides, respectively. Transformers capture them through local window partitions and adjacent window communications (\textit{e.g.}, window shifts and window expansions). However, like a coin with two sides, the success of CNNs and Transformers in preserving locality and continuity comes at the cost of their ability to capture long-range dependencies. Both of them only consider a limited region of the input image at a time due to their localized kernels or windows, making them challenging to model relationships that span across larger sections of the image. Therefore, it is highly expected to develop a method that is able to capture long-range dependencies while well preserving locality and continuity inherent in 2D natural images.

Mamba~\cite{mamba, mamba2}, a novel selective State Space Model~\cite{SSM}, has garnered significant attention due to its promising performance in long sequence modeling while maintaining nearly linear complexity. As Mamba's core algorithm, Selective Scan Operation (SSO), is inherently designed for 1D sequences, it can not be directly applicable to processing 2D images. To address the problem, Mamba-based restoration methods typically involve a 3-step pipeline: i) flattening 2D image into multiple 1D sequences along rows and columns; ii) processing each sequence independently using SSO; and iii) aggregating the processed sequences to form the output 2D image. However, such a paradigm still faces two demerits when processing images. First, when transforming image into sequences, it disrupts the locality and continuity inherent in image, as illustrated in~\cref{Figure:observation}(a)-(c). Second, it generally aggregates processed sequences via pixel-wise summation, overlooking the distinct contexts among sequences unfolded through totally different ways.

In this work, we present a novel locality- and continuity-preserving Mamba for Image Restoration (MaIR), which consists of Nested S-shaped Scanning strategy (NSS) and Sequence Shuffle Attention block (SSA). Specifically, NSS preserves the locality through stripe-based scanning region, and the continuity via the S-shaped scanning path with shift-stripe mechanism. SSA aggregates the processed sequences by calculating attention weights within corresponding channels of sequences. Thanks to corporation of NSS and SSA, MaIR enjoys the following merits. Firstly, MaIR involves a cost-free solution to preserve the locality and continuity inherent in natural images, ensuring structural coherence and avoiding computational overhead. Secondly, MaIR captures complex dependencies across distinct sequences, facilitating to leverage complementary information from both forward and reversed rows and columns.

To summarize, the contributions and innovations of this work are as below:
\begin{itemize}
	\item In this work, we present MaIR, an approach that efficiently captures long-range dependencies while preserving the locality and continuity inherent in natural images.
	\item For Mamba, we introduce NSS, a cost-free solution to preserve locality and continuity, and SSA, a module to capture dependencies across distinct sequences.
	\item MaIR obtains state-of-the-art performance on four tasks across 14 benchmarks comparing with 40 baselines.
\end{itemize}

\section{Related Works}
\label{Sec:RelatedWorks}
In this section, we will briefly review related works in image restoration and vision Mamba.

\subsection{Image Restoration}
According to the focus of this paper, existing methods can be classified into three categories, \textit{i.e.}, CNN-, Transformer- and Mamba-based methods. We will introduce the first two categories here, while the last one is detailed in~\cref{Sec:SSM}.

\textbf{CNN-based Method:}
Benefiting from the ability of capturing locality and continuity in natural images, CNN-based methods have achieved promising results in various tasks of image restoration, such as image super-resolution~\cite{EDSR,RCAN,SAN,HAN,LAPAR}, image denoising~\cite{DnCNN,FDD,RDN,clear,AirNet} and image deblurring~\cite{gopro,gopro,srn,dbgan}. However, since their localized receptive fields, CNNs are inherently limited in capturing long-range dependencies.

\textbf{Transformer-based Method:} Transformers are theoretically capable of capturing the global dependencies~\cite{CODE, Restormer}. However, to avoid impractical quadratic complexity on images, existing methods~\cite{swin, SwinIR, hat} tend to partition the local regions of input image into different windows, and calculate attentions within or across the windows. For instance, SwinIR~\cite{SwinIR} computes attentions within local windows and shifts these windows between layers. HAT~\cite{hat} divides images into overlapping windows to enhance the interaction between neighbor windows. Although these methods have ensured structural coherence (\textit{i.e.}, locality and continuity) of natural images and avoided computational overhead, they fell into another dilemma of failing to fully capture long-range dependencies due to their limited window sizes.

\subsection{Vision Mamba}
\label{Sec:SSM}
Due to Mamba's demonstrated superiority in long-sequence modeling~\cite{SSM, s5, mimo}, some studies have introduced it into high-~\cite{vmamba, vim, localmamba} and low-level~\cite{MambaIR, cumamba, uvmnet} vision tasks. To enable SSO to process images, these methods~\cite{vmamba,vim} tend to flatten 2D images into multiple 1D sequences along the different directions. For instance, Vmamba~\cite{vmamba} proposes cross-scan strategy which flattens input images along rows and columns. However, existing scanning strategies disrupt structure coherence which is essential for image restoration. Recently, some Mamba-based restoration methods have begun to recognize the importance of structure coherence, and tend to introduce extra coherence-preserving modules. For instance, MambaIR~\cite{MambaIR} and UVM-Net~\cite{uvmnet} enhances locality through additional CNN layers, but introduces extra computational costs. Although some other studies~\cite{localmamba, zigma} devote to designing scanning strategy to preserve locality and continuity, most of them can only preserve one of them. In contrast, MaIR provides a cost-free solution to preserve both locality and continuity.

\begin{figure*}[t!]
\begin{center}
\includegraphics[width=2.0\columnwidth]{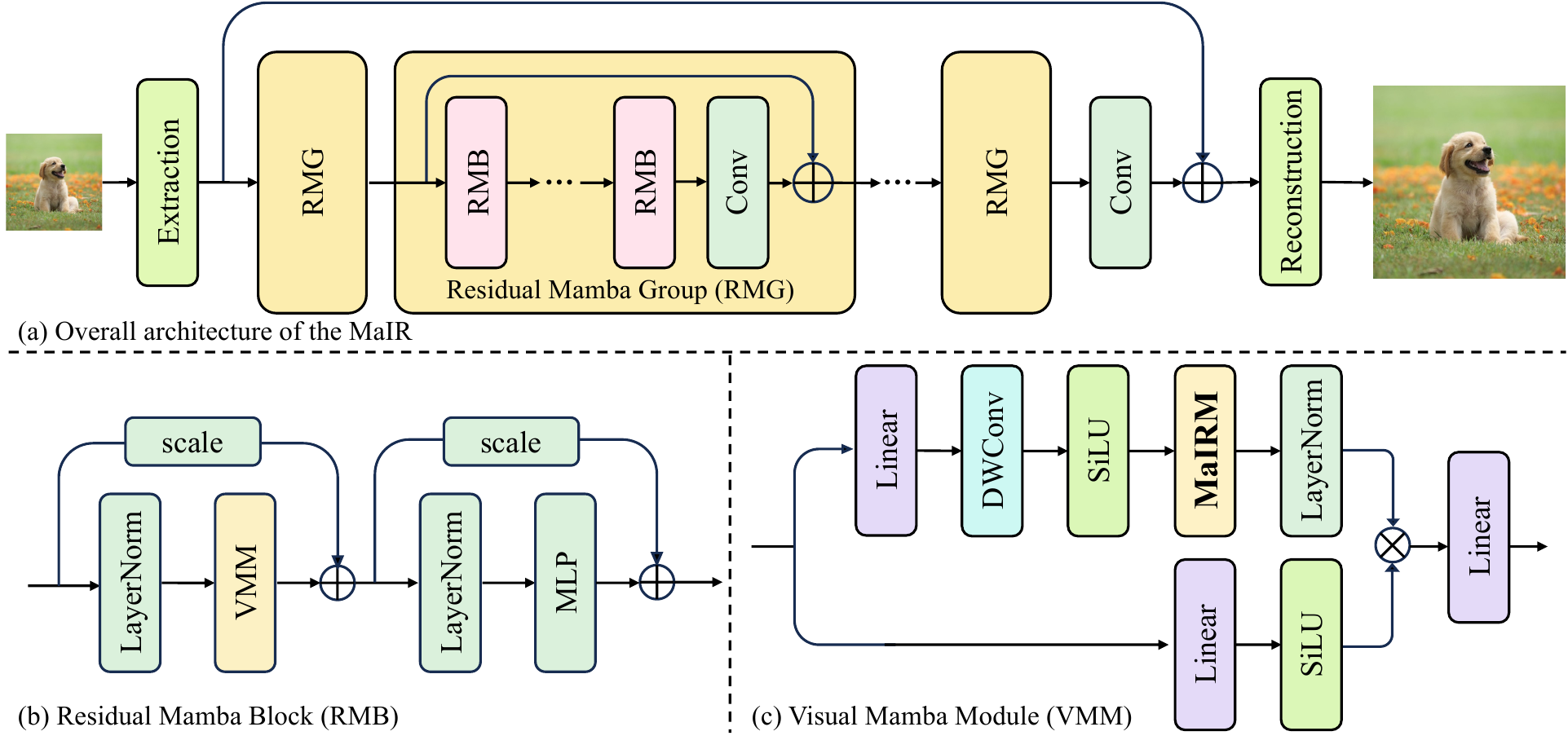}
\end{center}
\vspace{-1.2em}
\caption{Illustrations of MaIR. (a) The overall architecture of MaIR, highlighting its core component, Residual Mamba Group (RMG). RMG is primarily composed of (b) Residual Mamba Block (RMB), in which (c) Visual Mamba Module (VMM) plays a pivotal role.}
\label{Figure:architecture}
\end{figure*}

\begin{figure}[t!]
\begin{center}
  \includegraphics[width=1.0\columnwidth]{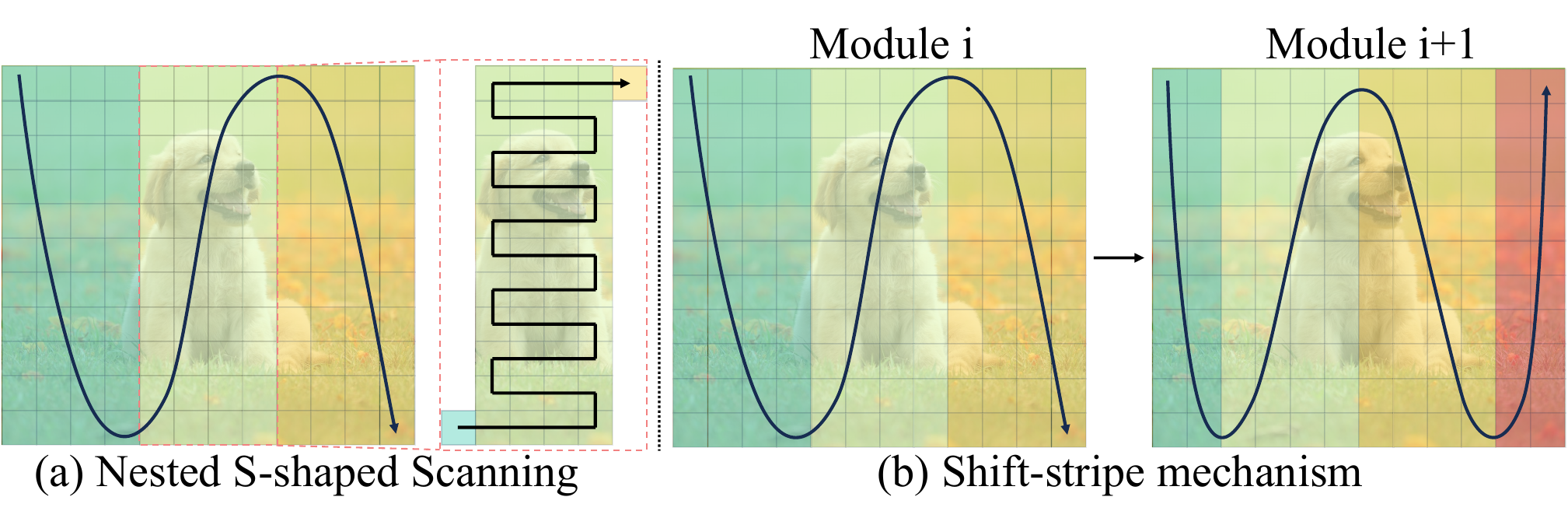}
\end{center}
\vspace{-2em}   
\caption{Illustrations of (a) Nested S-shaped Scanning strategy (NSS) and (b) shift-stripe mechanism.}
\label{Figure:nsss}
\vspace{-1.5em}   
\end{figure}

\section{Methods}
\label{Sec:Methods}
In this section, we first introduce the overall architecture of our MaIR, and then elaborate on NSS and SSA assembled in MaIR Module (MaIRM).

\subsection{Overall Architecture}
\textbf{Network Structure:} Following previous works~\cite{SwinIR,MambaIR}, MaIR is built up with three stages, namely, shallow feature extraction stage, deep feature extraction stage and reconstruction stage. Specifically, in the shallow feature extraction stage, for a given degraded image $x\in\mathcal{R}^{3\times H\times W}$, we first employ a convolution layer to extract shallow feature $F_S\in \mathcal{R}^{C\times H\times W}$, where $H$ and $W$ represent the height and width of $x$, and $C$ is the number of channels. After that, $F_S$ is fed to the deep feature extraction stage to produce deep feature $F_D\in \mathcal{R}^{C\times H\times W}$. As illustrated in~\cref{Figure:architecture}, the deep feature extraction stage is stacked by multiple Residual Mamba Groups (RMGs), where each RMG consists of several Residual Mamba Blocks (RMBs). Within each RMB, a Visual Mamba Module (VMM) is introduced to capture long-range dependencies, which is further composed of our proposed MaIRM. Finally, we reconstruct the high-quality image based on $F_S$ and $F_D$. Specifically, for image super-resolution, we introduce a pixel-shuffle layer $U_{ps}(\cdot)$ and a $3\times 3$ convolution layer $\Phi_{3\times 3}(\cdot)$ to reconstruct the high-resolution image $y' = \Phi_{3\times 3}(U_{ps}(F_{S} + F_{D}))$. For tasks that do not require upsampling (\textit{e.g.}, denoising, deblurring and dehazing), we employ single convolution layer with residual connection to construct high-quality result, which can be formulated as $y' = \Phi_{3\times 3}(F_{S} + F_{D}) + x$.

\textbf{Loss Function:}
For image super-resolution, we use $L_1$ loss to optimize the network following~\cite{SRFormer,SwinIR,MambaIR}, which can be formulated as 
\begin{equation*}
\mathcal{L}= \|y - y'\|_1,
\end{equation*}
where $y$ is the target image. For image denoising, deblurring and dehazing, we adopt Charbonnier loss, \textit{i.e.},
\begin{equation*}
\mathcal{L}= \sqrt{\|y - y'\|^2 + \epsilon^2},
\end{equation*}
where $\epsilon$ is a hyper-parameter and set to $10^{-3}$ empirically.


\subsection{MaIR Module}
As elaborated above, MaIRM serves as the core module of MaIR, which involves a three-step pipeline. To be specific, MaIRM first flattens 2D features into four 1D sequences through NSS along four distinct directions following~\cite{vmamba}. Then, MaIRM employs SSO to capture long-range dependencies. Finally, MaIRM aggregates processed sequences through SSA to form outputs. Mathematically, for input feature $F_{i,j}$, output feature $F^{M}_{i,j}$ can be formulated as 
\begin{equation*}
\begin{aligned}
	F^{M}_{i,j} &= M_{i,j}(F_{i,j}),\\
	&= \Phi_{i,j}^{SSA}(\Phi^{SSO}_{i,j}(\Phi_{i,j}^{NSS}(F_{i,j}))), \\
\end{aligned}
\end{equation*}
where $M_{i,j}(\cdot)$, $\Phi_{i,j}^{NSS}(\cdot)$, $\Phi_{i,j}^{SSO}(\cdot)$ and $\Phi_{i,j}^{SSA}(\cdot)$ are MaIRM, NSS, SSO and SSA in the $j$-th RMB of the $i$-th RMG, respectively.

\textbf{NSS:} 
NSS is designed to extract locality- and continuity-preserving sequences from input features. Motivated by the observation illustrated in~\cref{Figure:observation}, one could find that i) LocalMamba~\cite{localmamba} preserves locality through restricted scanning region, and ii) Zigma~\cite{zigma} preserves continuity through S-shaped scanning path. Thus, as shown in~\cref{Figure:nsss}(a), we design the nested S-shaped scanning strategy, which divides features into multiple non-overlapping stripes and uses S-shaped scanning path within and across stripes to maintain both locality and continuity. To better leverage spatial information, we extract sequences with four different scanning directions: top-left to bottom-right, bottom-right to top-left, top-right to bottom-left, and bottom-left to top-right, following previous works~\cite{vmamba,MambaIR}.

Besides, NSS includes shift-stipe mechanism to preserve locality and continuity on the boundary regions between adjacent stripes. As depicted in Fig.~\ref{Figure:nsss}(b), for two successive modules, the first module partitions features into multiple non-overlapping stripes with stripe width $w_s$. For the second module, we employ the shift-stripe operation, and set the first and last stripe widths as $\frac{w_s}{2}$ and others' width as $w_s$. Consequently, the boundary regions in the previous module will be fully covered by a single stripe in this module. 

\begin{figure*}[t]
\begin{center}
  \includegraphics[width=2.0\columnwidth]{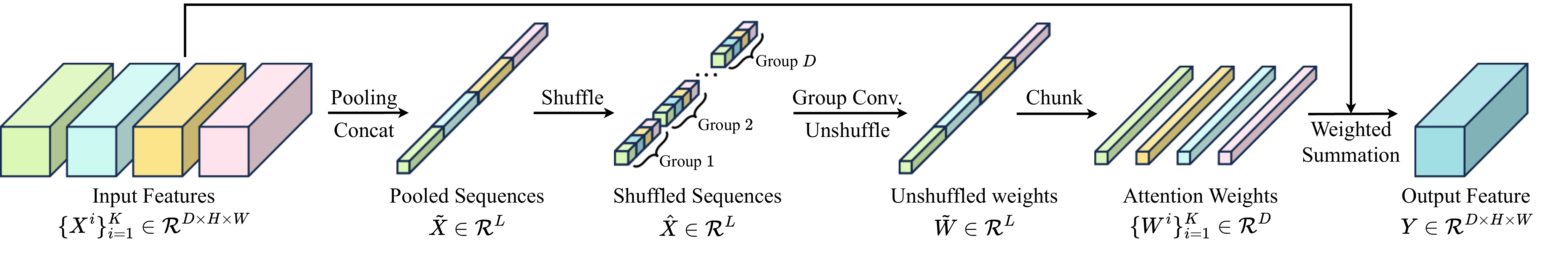}
\end{center}
\vspace{-2em}   
\caption{Illustration of the Sequence Shuffle Attention (SSA). The input features $\{X^i\}_{i=1}^K \in \mathcal{R}^{D \times H \times W}$ are first pooled and concatenated to form $\tilde{X} \in \mathcal{R}^{L}$, where $L = K \times D$. This sequence undergoes the sequence shuffle operation and results in shuffled sequences $\hat{X} \in \mathcal{R}^{L}$, whose channels are split by $D$ group. Then, group convolution and sequence unshuffle operation are applied, producing unshuffled weights $\tilde{W} \in \mathcal{R}^{L}$, which are further chunked and reshaped into attention weights $\{W^i\}_{i=1}^K \in \mathcal{R}^{D}$. Finally, the output feature $Y \in \mathcal{R}^{D \times H \times W}$ is computed by performing a weighted summation of the input features using the attention weights.}
\vspace{-1em}   
\label{Figure:SSA}
\end{figure*}

\textbf{SSA:} SSA aggregates the processed sequences by calculating attentions within corresponding channels. This design enables it to capture complex dependencies across distinct sequences, thus better leveraging complementary information from different scanning directions.
As shown in~\cref{Figure:SSA}, supposing sequence number $K=4$, for SSO-processed sequences $\lbrace X^i \rbrace_{i=1}^{4}$, we first apply spatial average pooling $\Phi_{AP}(\cdot)$ to reduce the computational cost, and then concatenate as
\begin{equation*}
\begin{aligned}
\tilde{X}&=\Phi_{cat}(\Phi_{AP}(\lbrace X^i \rbrace_{i=1}^{4}))\\
 &= [x^{1}_1, \cdots, x^{1}_D, x^{2}_1, \cdots, x^{2}_D, x^{3}_1, \cdots, x^{3}_D, x^{4}_1, \cdots, x^{4}_D],
\end{aligned}
\end{equation*}
where $x^k_d$ is the pooled feature in $d$-th channel of $k$-th sequence, and $D$ is the number of channel in MaIRM. Then, we employ sequence shuffle operation $\Phi_{ss}(\cdot)$ to rearrange features into 
\begin{equation*}
\begin{aligned}
\hat{X}&=\Phi_{ss}(X)\\
&=[x^{1}_1, x^{2}_1, x^{3}_1, x^{4}_1, x^{1}_2, x^{2}_2, x^{3}_2, x^{4}_2, \cdots, x^{1}_D, x^{2}_D, x^{3}_D, x^{4}_D].
\end{aligned}
\end{equation*}
After that, we employ group convolution $\Phi_{g}(\cdot)$ with group size four to obtain the channel-wise attention weights and unshuffle the weights back to their original order, \textit{i.e.},
\begin{equation*}
\begin{aligned}
\tilde{W} &= \Phi_{su}(\Phi_{g}(\hat{X}))\\
&= [w^{1}_1, \cdots, w^{1}_D, w^{2}_1, \cdots, w^{2}_D, w^{3}_1, \cdots, w^{3}_D, w^{4}_1, \cdots, w^{4}_D],
\end{aligned}
\end{equation*}
where $\Phi_{su}(\cdot)$ is sequence unshuffle operation. The unshuffled weights $\tilde{W}$ are chunked as $
\lbrace W^i \rbrace_{i=1}^{4} = \Phi_{chunk}(\tilde{W}),
$
where $\Phi_{chunk}(\cdot)$ refers to the chunk operation. Finally, we adopt weight summation based on $\lbrace W^i \rbrace_{i=1}^{4}$ to generate the output, which can be formulated as: 
\begin{equation*}
\begin{aligned}
Y &= \sum_{i=1}^{K=4} W^i * X^i,
\end{aligned}
\end{equation*}
and $Y$ is the output sequence of SSA. 

\section{Experiments}
\label{Sec:Experiments}
In this section, we evaluate our MaIR on four representative image restoration tasks, \textit{i.e.}, image super-resolution, image denoising, image deblurring, and image dehazing. In the following, we will first show quantitative results, and then conduct analysis studies to verify the reasonability. Experimental settings will be presented in the supplementary materials.

\begin{table*}[t]
    \centering
    \normalsize
    \caption{Quantitative results on classic image super-resolution. The best and second best results are in \textcolor{red}{red} and \textcolor{blue}{blue}.}
	
    \begin{tabular}{l | c | c c | c c | c c | c c | c c}
	\toprule
         \multirow{2}{*}{Methods} & \multirow{2}{*}{Scale} & \multicolumn{2}{c|}{Set5} & \multicolumn{2}{c|}{Set14} & \multicolumn{2}{c|}{B100} & \multicolumn{2}{c}{Urban100} & \multicolumn{2}{c}{Manga109}\\
            & & PSNR & SSIM & PSNR & SSIM & PSNR & SSIM & PSNR & SSIM & PSNR & SSIM \\
         \midrule
		 SAN~\cite{SAN}
          & $\times 2$ & 38.31  & 0.9620 & 34.07 & 0.9213 & 32.42 & 0.9028 & 33.10 & 0.9370 & 39.32 & 0.9792\\
		 HAN~\cite{HAN}
          & $\times 2$ & 38.27 & 0.9614  & 34.16 & 0.9217 & 32.41 & 0.9027 & 33.35 & 0.9385 & 39.46 & 0.9785 \\
		 IGNN~\cite{IGNN}
          & $\times 2$ & 38.24  & 0.9613  & 34.07  & 0.9217 & 32.41 & 0.9025 & 33.23 & 0.9383 & 39.35 & 0.9786 \\
		 NLSA~\cite{NLSA}    
          & $\times 2$ & 38.34 & 0.9618 & 34.08 & 0.9231 & 32.43 & 0.9027 & 33.42 & 0.9394 & 39.59 & 0.9789\\
          ELAN~\cite{ELAN}    
          & $\times 2$ & 38.36 & 0.9620 & 34.20 & 0.9228 & 32.45 & 0.9030 & 33.44 & 0.9391 & 39.62 & 0.9793\\
		 IPT~\cite{IPT}  
          & $\times 2$ & 38.37 & - & 34.43 & - & 32.48 & - & 33.76 & - & - & - \\
		 SwinIR~\cite{SwinIR}     
          & $\times 2$ & 38.42 & 0.9623 & 34.46 & 0.9250 & 32.53 & 0.9041 & 33.81 & 0.9427 & 39.92 & 0.9797\\
		 SRFormer~\cite{SRFormer}     
          & $\times 2$ & 38.51 & 0.9627 & 34.44 & 0.9253 & 32.57 & 0.9046 & 34.09 & 0.9449 & 40.07 & 0.9802 \\
		 MambaIR~\cite{MambaIR}
          & $\times 2$ & \textcolor{blue}{38.57} & 0.9627 & 34.67 & 0.9261 & 32.58 & 0.9048 & 34.15 & 0.9446 & 40.28 & 0.9806\\
		 MaIR
          & $\times 2$  & 38.56 & \textcolor{blue}{0.9628} & \textcolor{blue}{34.75} & \textcolor{blue}{0.9268} & \textcolor{blue}{32.59} & \textcolor{blue}{0.9049} & \textcolor{blue}{34.19} & \textcolor{blue}{0.9451}  & \textcolor{blue}{40.30}  & \textcolor{blue}{0.9807}\\
		 MaIR+
          & $\times 2$  & \textcolor{red}{38.62} & \textcolor{red}{0.9630} & \textcolor{red}{34.82} & \textcolor{red}{0.9272} & \textcolor{red}{32.62} & \textcolor{red}{0.9053} & \textcolor{red}{34.38} & \textcolor{red}{0.9462}  & \textcolor{red}{40.48}  & \textcolor{red}{0.9811}\\
         \midrule
		 SAN~\cite{SAN}
          & $\times 3$ & 34.75 & 0.9300 & 30.59 & 0.8476 & 29.33 & 0.8112 & 28.93 & 0.8671 & 34.30 & 0.9494\\
		 HAN~\cite{HAN}
          & $\times 3$ & 34.75 & 0.9299 & 30.67 & 0.8483 & 29.32 & 0.8110 & 29.10 & 0.8705 & 34.48 & 0.9500 \\
		 IGNN~\cite{IGNN}
          & $\times 3$ & 34.72 & 0.9298 & 30.66 & 0.8484 & 29.31 & 0.8105 & 29.03 & 0.8696 & 34.39 & 0.9496 \\
		 NLSA~\cite{NLSA}    
          & $\times 3$ & 34.85 & 0.9306 & 30.70 & 0.8485 & 29.34 & 0.8117 & 29.25 & 0.8726 & 34.57 & 0.9508\\
          ELAN~\cite{ELAN}    
          & $\times 3$ & 34.90 & 0.9313 & 30.80 & 0.8504 & 29.38 & 0.8124 & 29.32 & 0.8745 & 34.73 & 0.9517\\
		 IPT~\cite{IPT}  
          & $\times 3$ & 34.81 & - & 30.85 & - & 29.38 & - & 29.49 & - & - & - \\
		 SwinIR~\cite{SwinIR}     
          & $\times 3$ & 34.97 & 0.9318 & 30.93 & 0.8534 & 29.46 & 0.8145 & 29.45 & 0.8826 & 35.12 & 0.9537\\
		 SRFormer~\cite{SRFormer}     
          & $\times 3$ & 35.02 & 0.9323 & 30.94 & 0.8540 & 29.48 & 0.8156 & 30.04 & \textcolor{blue}{0.8865} & 35.26 & 0.9543\\
		 MambaIR~\cite{MambaIR}
          & $\times 3$ & 35.08 & 0.9323 & 30.99 & 0.8536 & 29.51 & 0.8157 & 29.93 & 0.8841 & 35.43 & 0.9546\\
		 MaIR
          & $\times 3$  & \textcolor{blue}{35.10} & \textcolor{blue}{0.9324}  & \textcolor{blue}{31.05} & \textcolor{blue}{0.8541} & \textcolor{blue}{29.51}  & \textcolor{blue}{0.8160}  & \textcolor{blue}{30.05}  & 0.8863  & \textcolor{blue}{35.44}  & \textcolor{blue}{0.9547} \\
		 MaIR+
          & $\times 3$  & \textcolor{red}{35.15} & \textcolor{red}{0.9328}  & \textcolor{red}{31.12} & \textcolor{red}{0.8550} & \textcolor{red}{29.56}  & \textcolor{red}{0.8167}  & \textcolor{red}{30.24}  & \textcolor{red}{0.8881}  & \textcolor{red}{35.67}  & \textcolor{red}{0.9556} \\

         \midrule
		 SAN~\cite{SAN}
          & $\times 4$ & 32.64 & 0.9003 & 28.92 & 0.7888 & 27.78 & 0.7436 & 26.79 & 0.8068 & 31.18 & 0.9169\\
		 HAN~\cite{HAN}
          & $\times 4$ & 32.64 & 0.9002 & 28.90 & 0.7890 & 27.80 & 0.7442 & 26.85 & 0.8094 & 31.42 & 0.9177 \\
		 IGNN~\cite{IGNN}
          & $\times 4$ & 32.57 & 0.8998 & 28.85 & 0.7891 & 27.77 & 0.7434 & 26.84 & 0.8090 & 31.28 & 0.9182 \\
		 NLSA~\cite{NLSA}    
          & $\times 4$ & 32.59 & 0.9000 & 28.87 & 0.7891 & 27.78 & 0.7444 & 26.96 & 0.8109 & 31.27 & 0.9184 \\
          ELAN~\cite{ELAN}    
          & $\times 4$ & 32.75 & 0.9022 & 28.96 & 0.7914 & 27.83 & 0.7459 & 27.13 & 0.8167 & 31.68 & 0.9226 \\
		 IPT~\cite{IPT}  
          & $\times 4$ & 32.64 & - & 29.01 & - & 27.82 & - & 27.26 & - & - & - \\
		 SwinIR~\cite{SwinIR}     
          & $\times 4$ & 32.92 & 0.9044 & 29.09 & 0.7950 & 27.92 & 0.7489 & 27.45 & 0.8254 & 32.03 & 0.9260\\
		 SRFormer~\cite{SRFormer}     
          & $\times 4$ & 32.93 & 0.9041 & 29.08 & 0.7953 & 27.94 & 0.7502 & 27.68 & \textcolor{blue}{0.8311} & 32.21 & 0.9271\\
		 MambaIR~\cite{MambaIR}
          & $\times 4$ & \textcolor{blue}{33.03} & \textcolor{blue}{0.9046} & \textcolor{blue}{29.20} & \textcolor{blue}{0.7961} & 27.98 & 0.7503 & 27.68 & 0.8287 & 32.32 & 0.9272\\
		 MaIR
          & $\times 4$  &  32.93 & 0.9045  & 29.20  & 0.7958  &  \textcolor{blue}{27.98} & \textcolor{blue}{0.7507} & \textcolor{blue}{27.71} & 0.8305 & \textcolor{blue}{32.46} & \textcolor{blue}{0.9284} \\
          MaIR+
          & $\times 4$  & \textcolor{red}{33.14} & \textcolor{red}{0.9058}  & \textcolor{red}{29.28}  & \textcolor{red}{0.7974}  & \textcolor{red}{28.02} & \textcolor{red}{0.7516} & \textcolor{red}{27.89} & \textcolor{red}{0.8336} & \textcolor{red}{32.66} & \textcolor{red}{0.9297} \\
		 \bottomrule
    \end{tabular}
    \label{Tab:ClassicSR}
    \vspace{-1em}
\end{table*}

\begin{table*}[t]
    \centering
    \footnotesize
    \caption{Quantitative results on lightweight image super-resolution. The best and second best results are in \textcolor{red}{red} and \textcolor{blue}{blue}.}
    \resizebox{1.0\textwidth}{!}{
    \begin{tabular}{l | c | c | c | c c | c c | c c | c c | c c}
	\toprule
         \multirow{2}{*}{Methods} & \multirow{2}{*}{Scale} & \multirow{2}{*}{Params} & \multirow{2}{*}{MACs} & \multicolumn{2}{c|}{Set5} & \multicolumn{2}{c|}{Set14} & \multicolumn{2}{c|}{B100} & \multicolumn{2}{c}{Urban100} & \multicolumn{2}{c}{Manga109}\\
            & &  &  & PSNR & SSIM & PSNR & SSIM & PSNR & SSIM & PSNR & SSIM & PSNR & SSIM \\
         \midrule
	   CARN~\cite{CARN}    
          & $\times 2$ & 1,592K & 222.8G & 37.76 & 0.9590 & 33.52 & 0.9166 & 32.09 & 0.8978 & 31.92 & 0.9256 & 38.36 & 0.9765\\
		 IMDN~\cite{IMDN}
          & $\times 2$ & 694K & 158.8G & 38.00 & 0.9605 & 33.63 & 0.9177 & 32.19 & 0.8996 & 32.17 & 0.9283 & 38.88 & 0.9774\\
		 LAPAR-A~\cite{LAPAR}
          & $\times 2$ & 548K & 171.0G & 38.01 & 0.9605 & 33.62 & 0.9183 & 32.19 & 0.8999 & 32.10 & 0.9283 & 38.67 & 0.9772\\
		 LatticeNet~\cite{Latticenet}
          & $\times 2$ & 756K & 169.5G & 38.15 & 0.9610 & 33.78 & 0.9193 & 32.25 & 0.9005 & 32.43 & 0.9302 & - & - \\
		 SwinIR~\cite{SwinIR}     
          & $\times 2$ & 910K & 122.2G & 38.14 & \textcolor{blue}{0.9611} & 33.86 & 0.9206 & 32.31 & 0.9012 & 32.76 & 0.9340 & 39.12 & \textcolor{red}{0.9783}\\
		 MambaIR-Tiny~\cite{MambaIR}
          & $\times 2$ & 905K & 167.1G & 38.13 & 0.9610 & \textcolor{blue}{33.95} & 0.9208 & 32.31 & 0.9013 & 32.85  & 0.9349 & 39.20 & \textcolor{blue}{0.9782}\\
		 MaIR-Tiny
          & $\times 2$  & 878K & 207.8G & \textcolor{blue}{38.18} & 0.9610 & 33.89 & 0.9209 & 32.31 & 0.9013 & 32.89 & 0.9346 & 39.22 & 0.9778\\
		 MambaIR-Small~\cite{MambaIR}
          & $\times 2$ & 1,363K & 567.5G & 38.16 & 0.9610 & \textcolor{red}{34.00} & \textcolor{red}{0.9212} & \textcolor{red}{32.34} & \textcolor{red}{0.9017} & \textcolor{blue}{32.92} & \textcolor{blue}{0.9356} & \textcolor{blue}{39.31} & 0.9779 \\
		 MaIR-Small
          & $\times 2$ & 1,355K & 542.0G  & \textcolor{red}{38.20} & \textcolor{red}{0.9611} & 33.91 & \textcolor{blue}{0.9209} & \textcolor{blue}{32.34} & \textcolor{blue}{0.9016} & \textcolor{red}{32.97} & \textcolor{red}{0.9359} & \textcolor{red}{39.32} & 0.9779 \\

         \midrule
	   CARN~\cite{CARN}    
          & $\times 3$ & 1,592K & 111.8G & 34.29 & 0.9255 & 30.29 & 0.8407 & 29.06 & 0.8034 & 28.06 & 0.8493 & 33.50 & 0.9440\\
		 IMDN~\cite{IMDN}
          & $\times 3$ & 703K & 71.5G & 34.36 & 0.9270 & 30.32 & 0.8417 & 29.09 & 0.8046 & 28.17 & 0.8519 & 33.61 & 0.9445\\
		 LAPAR-A~\cite{LAPAR}
          & $\times 3$ & 544K & 144.0G & 34.36 & 0.9267 & 30.34 & 0.8421 & 29.11 & 0.8054 & 28.15 & 0.8523 & 33.51 & 0.9441\\
		 LatticeNet~\cite{Latticenet}
          & $\times 3$ & 765K & 76.3G & 34.53 & 0.9281 & 30.39 & 0.8424 & 29.15 & 0.8059 & 28.33 & 0.8538 & - & - \\
		 SwinIR~\cite{SwinIR}     
          & $\times 3$ & 918K & 55.4G & 34.62 & 0.9289 & 30.54 & 0.8463 & 29.20 & 0.8082 & 28.66 & 0.8624 & 33.98 & 0.9478 \\
		 MambaIR-Tiny~\cite{MambaIR}
          & $\times 3$ & 913K & 74.5G & 34.63 & 0.9288 & 30.54 & 0.8459 & 29.23 & 0.8084 & 28.70 & 0.8631 & 34.12 & 0.9479\\
		 MaIR-Tiny
          & $\times 3$ & 886K & 93.0G & 34.68 & 0.9292 & 30.54 & 0.8461 & 29.25 & 0.8088 & 28.83 & 0.8651 & 34.21 & 0.9484\\
		 MambaIR-Small~\cite{MambaIR}
          & $\times 3$ & 1,371K & 252.7G & \textcolor{blue}{34.72} & \textcolor{blue}{0.9296} & \textcolor{red}{30.63} & \textcolor{blue}{0.8475} & \textcolor{blue}{29.29} & \textcolor{blue}{0.8099} & \textcolor{red}{29.00} & \textcolor{red}{0.8689} & \textcolor{blue}{34.39} & \textcolor{blue}{0.9495}\\
		 MaIR-Small
          & $\times 3$ & 1,363K & 241.4G & \textcolor{red}{34.75} & \textcolor{red}{0.9300} & \textcolor{blue}{30.63} & \textcolor{red}{0.8479} & \textcolor{red}{29.29} & \textcolor{red}{0.8103} & \textcolor{blue}{28.92} & \textcolor{blue}{0.8676} & \textcolor{red}{34.46} & \textcolor{red}{0.9497} \\

         \midrule
	   CARN~\cite{CARN}    
          & $\times 4$ & 1,592K & 90.9G & 32.13 & 0.8937 & 28.60 & 0.7806 & 27.58 & 0.7349 & 26.07 & 0.7837 & 30.47 & 0.9084\\
		 IMDN~\cite{IMDN}
          & $\times 4$ & 715K & 40.9G & 32.21 & 0.8948 & 28.58 & 0.7811 & 27.56 & 0.7353 & 26.04 & 0.7838 & 30.45 & 0.9075\\
		 LAPAR-A~\cite{LAPAR}
          & $\times 4$ & 659K & 94.0G & 32.15 & 0.8944 & 28.61 & 0.7818 & 27.61 & 0.7366 & 26.14 & 0.7871 & 30.42 & 0.9074\\
		 LatticeNet~\cite{Latticenet}
          & $\times 4$ & 777K & 43.6G & 32.30 & 0.8962 & 28.68 & 0.7830 & 27.62 & 0.7367 & 26.25 & 0.7873 & - & - \\
		 SwinIR~\cite{SwinIR}     
          & $\times 4$ & 930K & 31.8G & 32.44 & 0.8976 & 28.77 & 0.7858 & 27.69 & 0.7406 & 26.48 & 0.7980 & 30.92 & 0.9151\\
		 MambaIR-Tiny~\cite{MambaIR}
          & $\times 4$ & 924K & 42.3G & 32.42 & 0.8977 & 28.74 & 0.7847 & 27.68 & 0.7400 & 26.52 & 0.7983 & 30.94 & 0.9135\\
		 MaIR-Tiny
          & $\times 4$ & 897K & 53.1G & 32.48 & 0.8985 & 28.81 & 0.7864 & 27.71 & 0.7414 & 26.60 & 0.8013 & 31.13 & 0.9161 \\
		 MambaIR-Small~\cite{MambaIR}
          & $\times 4$ & 1,383K & 143.0G & \textcolor{blue}{32.51} & \textcolor{blue}{0.8993} & \textcolor{blue}{28.85} & \textcolor{blue}{0.7876} & \textcolor{blue}{27.75} & \textcolor{blue}{0.7423} & \textcolor{red}{26.75} & \textcolor{red}{0.8051} & \textcolor{blue}{31.26} & \textcolor{blue}{0.9175} \\
		 MaIR-Small
          & $\times 4$  & 1,374K & 136.6G & \textcolor{red}{32.62} & \textcolor{red}{0.8998} & \textcolor{red}{28.90} & \textcolor{red}{0.7882} & \textcolor{red}{27.77} & \textcolor{red}{0.7431} & \textcolor{blue}{26.73} & \textcolor{blue}{0.8049} & \textcolor{red}{31.34} & \textcolor{red}{0.9183} \\
    \bottomrule
    \end{tabular}
    }
    \label{Tab:LightweightSR}
    \vspace{-1em}
\end{table*}

\begin{figure*}[t]
\begin{center}
  \includegraphics[width=2.0\columnwidth]{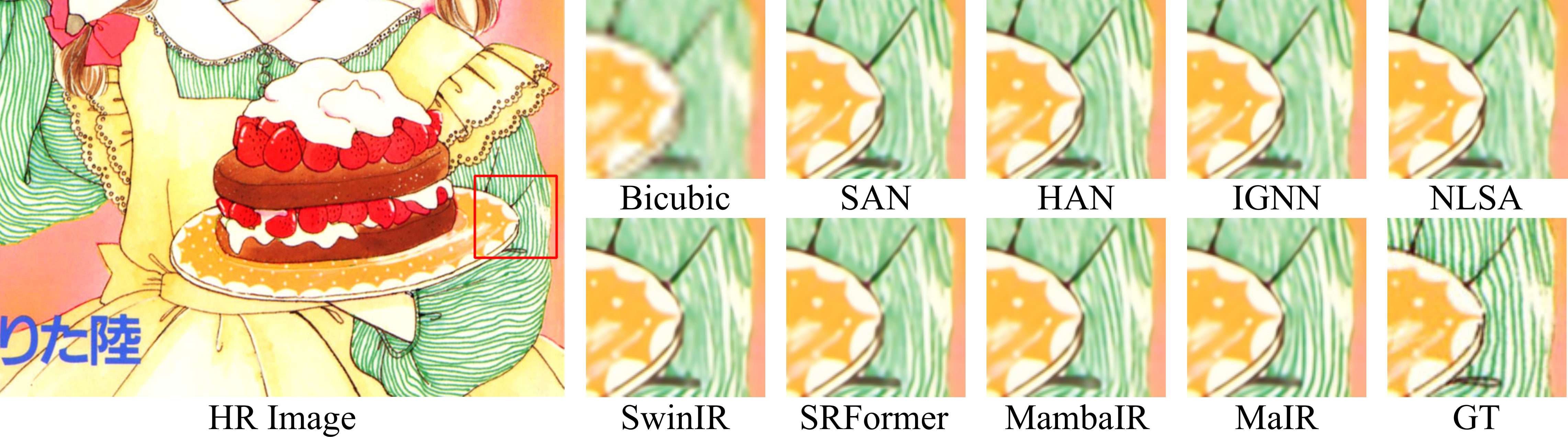}
\end{center}
\vspace{-1.5em}   
\caption{Visual comparison of $\times4$ image super-resolution results on the Manga109 dataset. MaIR demonstrates superior visual quality, particularly in preserving fine details and textures.}
\vspace{-1.5em}   
\label{Figure:sr}
\end{figure*}

\subsection{Results on Image Super-Resolution}
In this section, we conduct experiments on both classic and lightweight image super-resolution.

\textbf{Datasets:} Following previous works~\cite{SwinIR,MambaIR}, we employ DF2K (DIV2K~\cite{div2k}+Flickr2K~\cite{flickr2k}) as the training set for classic image super-resolution, and DIV2K as training set for lightweight image super-resolution. For evaluation, we employ the following five datasets as test sets, \textit{i.e.}, Set5~\cite{set5}, Set14~\cite{set14}, B100~\cite{BSD}, Urban100~\cite{urban100} and Manga109~\cite{manga109}. 
Following existing works~\cite{RCAN,IRCNN,SAN,HAN}, the low-resolution images are downsampled from the corresponding high-resolution images via bicubic interpolation. 

\textbf{Baselines:} We compare our method with 15 competitive baselines. Specifically, we adopt four CNN-based methods (\textit{i.e.}, SAN~\cite{SAN}, HAN~\cite{HAN}, IGNN~\cite{IGNN}, and NLSA~\cite{NLSA}), four transformer-based methods (\textit{i.e.}, ELAN~\cite{ELAN}, IPT~\cite{IPT}, SwinIR~\cite{SwinIR} and SRFormer~\cite{SRFormer}) and one Mamba-based method (\textit{i.e.,} MambaIR~\cite{MambaIR}) as the baselines for classic super-resolution. For lightweight super-resolution, four CNN-based methods (\textit{i.e.}, CARN~\cite{CARN}, IMDN~\cite{IMDN}, LAPAR~\cite{LAPAR}, LatticeNet~\cite{Latticenet}), two transformer-based methods (\textit{i.e.}, SwinIR~\cite{SwinIR} and SRFormer~\cite{SRFormer}) and one Mamba-based method (\textit{i.e.,} MambaIR~\cite{MambaIR}) are introduced into both quantitative and qualitative comparisons. Similar to MambaIR, which offers two versions for lightweight super-resolution, MaIR is also available in two configurations: MaIR-Tiny and MaIR-Small.

\textbf{Results:} For classic super-resolution, as shown in~\cref{Tab:ClassicSR,Figure:sr}, one could observe that MaIR achieves the best result in almost all quantitative comparisons. For instance, our method surpasses MambaIR~\cite{MambaIR} with 0.03dB$\sim$0.12dB in terms of PSNR on Urban100, and SRFormer with at most 0.04dB, 0.10dB and 0.25dB in terms of PSNR on B100, Urban100, and Manga109, respectively, which demonstrates the superiority of MaIR. For light-weight SR, MaIR also exhibits its advancement compared to baselines as reported in~\cref{Tab:LightweightSR}. Taking $\times4$ scale as examples, MaIR-Small surpasses MambaIR-Small by 0.08dB in terms of PSNR on Manga109 with fewer parameters and MACs. MaIR-Tiny outperforms MambaIR-Tiny and SwinIR by 0.08dB and 0.12dB in terms of PSNR on Urban100 with fewer parameters, which verifies both the efficiency and effectiveness of our proposed method.

\begin{table*}[h]
    \centering
    \normalsize
    \caption{Quantitative results on gaussian color image denoising. The best and second best results are in \textcolor{red}{red} and \textcolor{blue}{blue}.}
	
    \begin{tabular}{l | c c c | c c c | c c c | c c c}
	\toprule
         \multirow{2}{*}{Methods} & \multicolumn{3}{c|}{BSD68} & \multicolumn{3}{c|}{Kodak24} & \multicolumn{3}{c|}{McMaster} & \multicolumn{3}{c}{Urban100}\\
            & $\sigma$=15 & $\sigma$=25 & $\sigma$=50 & $\sigma$=15 & $\sigma$=25 & $\sigma$=50 & $\sigma$=15 & $\sigma$=25 & $\sigma$=50 & $\sigma$=15 & $\sigma$=25 & $\sigma$=50 \\
         \midrule
	   IRCNN~\cite{IRCNN}    
          & 33.86 & 31.16 & 27.86 & 34.69 & 32.18 & 28.93 & 34.58 & 32.18 & 28.91 & 33.78 & 31.20 & 27.70 \\
		 FFDNet~\cite{ffdnet}
          & 33.87 & 31.21 & 27.96 & 34.63 & 32.13 & 28.98 & 34.66 & 32.35 & 29.18 & 33.83 & 31.40 & 28.05\\
		 DnCNN~\cite{DnCNN}
          & 33.90 & 31.24 & 27.95 & 34.60 & 32.14 & 28.95 & 33.45 & 31.52 & 28.62 & 32.98 & 30.81 & 27.59\\
		 DRUNet~\cite{drunet}
          & 34.30 & 31.69 & 28.51 & 35.31 & 32.89 & 29.86 & 35.40 & 33.14 & 30.08 & 34.81 & 32.60 & 29.61\\
		 SwinIR~\cite{SwinIR}
          & 34.42 & 31.78 & 28.56 & 35.34 & 32.89 & 29.79 & 35.61 & 33.20 & 30.22 & 35.13 & 32.90 & 29.82\\
		 Restormer~\cite{Restormer}  
          & 34.40 & 31.79 & 28.60 & 35.47 & 33.04 & 30.01 & 35.61 & 33.34 & 30.30 & 35.13 & 32.96 & 30.02\\
          CODE~\cite{CODE}  
          & 34.33 & 31.69 & 28.47 & 35.32 & 32.88 & 29.82 & 35.38 & 33.11 & 30.03 & - & - & -\\
		 ART~\cite{art}
          & 34.46 & 31.84 & 28.63 & 35.39 & 32.95 & 29.87 & 35.68  & 33.41 & 30.31 & 35.29  & 33.14 & 30.19\\
     
		 MambaIR~\cite{MambaIR}
          & 34.43 & 31.80 & 28.61 & 35.34 & 32.91 & 29.85 & 35.62 & 33.35 & 30.31 & 35.17 & 32.99 & 30.07\\
          
		 MaIR
          & \textcolor{blue}{34.48} & \textcolor{blue}{31.86} & \textcolor{blue}{28.66} & \textcolor{blue}{35.53} & \textcolor{blue}{33.09} & \textcolor{blue}{30.04} & \textcolor{blue}{35.71} & \textcolor{blue}{33.44} & \textcolor{blue}{30.35} & \textcolor{blue}{35.35}  & \textcolor{blue}{33.22} & \textcolor{blue}{30.30}\\

		 MaIR+
          & \textcolor{red}{34.50}  & \textcolor{red}{31.88} & \textcolor{red}{28.69} & \textcolor{red}{35.56} & \textcolor{red}{33.13} & \textcolor{red}{30.08} & \textcolor{red}{35.74} & \textcolor{red}{33.48} & \textcolor{red}{30.39}  & \textcolor{red}{35.42}  & \textcolor{red}{33.30}  & \textcolor{red}{30.41} \\
         \bottomrule
    \end{tabular}
    \label{Tab:guassianDN}
\end{table*}

%

\begin{table*}[h!]
    \centering
    \normalsize
    \caption{Quantitative results on real image denoising. The best and second best results are in \textcolor{red}{red} and \textcolor{blue}{blue}.}
	
    \begin{tabular}{l |c c c c c  c  c }
	\toprule
         & {DeamNet~\cite{deamnet}} & {MPRNet~\cite{MPRNet}} & 
         {NBNet~\cite{nbnet}} & 
         {DAGL~\cite{dagl}} & {Uformer~\cite{uformer}} & {MambaIR~\cite{MambaIR}} & {MaIR} \\
         \midrule
	   PSNR    
          & 39.47 & 39.71 & 39.75 &38.94 & 39.89 & \textcolor{blue}{39.89} & \textcolor{red}{39.92} \\
	   SSIM    
          & 0.957 & 0.958 & 0.959 & 0.953 & 0.960 & \textcolor{blue}{0.960} & \textcolor{red}{0.960} \\
         \bottomrule
    \end{tabular}
    \label{Tab:realDN}
\end{table*}

\begin{figure*}[t!]
\begin{center}
  \includegraphics[width=2.0\columnwidth]{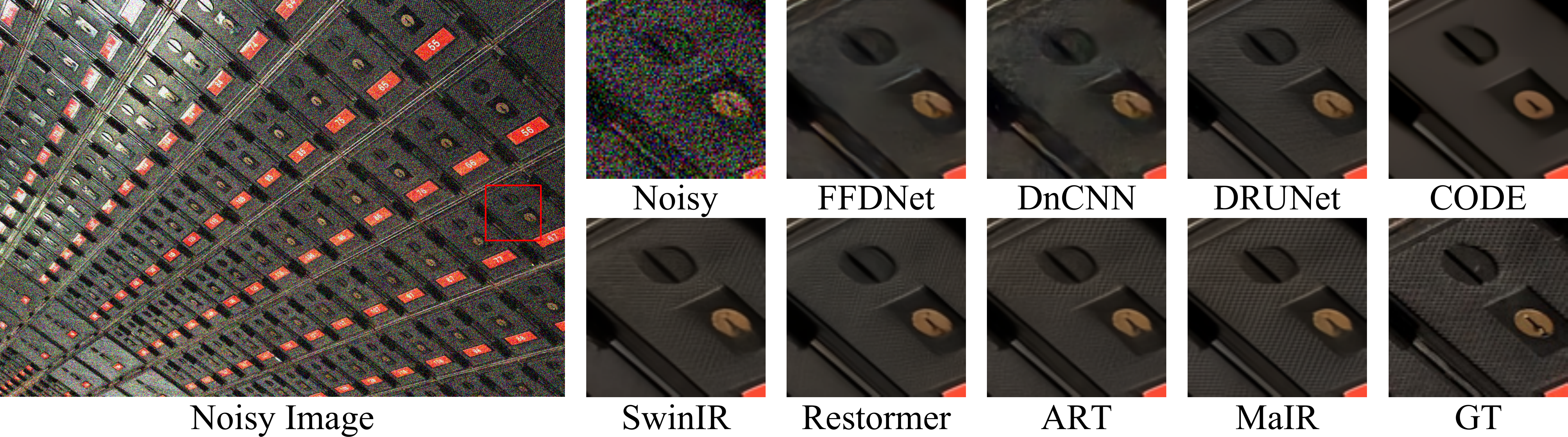}
\end{center}
\vspace{-1.5em}   
\caption{Visual comparison of image denoising results on the Urban100 dataset. MaIR effectively removes noise in the images and produces detailed textures that closely match the ground truth.}
\label{Figure:denoise}
\end{figure*}

\subsection{Results on Image Denoising}
In this section, we evaluate MaIR on both synthetic Gaussian noise and real-world noise.

\textbf{Datasets:} For synthetic noise removal, we train MaIR on DFWB, which consists of DIV2K, Flickr2K, Waterloo Exploration Dataset (WED)~\cite{WED} and BSD400~\cite{BSD}. For evaluation, we utilize BSD68~\cite{BSD}, Kodak24, McMaster~\cite{mcmaster}, and Urban100 as test set. Following~\cite{DnCNN,ffdnet,IRCNN,SwinIR}, we generate noisy images by manually adding white Gaussian noise to the clean images with three distinct noise levels, \textit{i.e.}, $\sigma=15, 25, 50$. For real-world image denoising, our model is trained and tested on the SIDD-Medium~\cite{sidd} dataset, which provides 320 high-resolution noisy-clean image pairs for training and additional 40 image pairs for test.

\textbf{Baselines:} We compare our MaIR with 14 representative methods. To be specific, we adopt four CNN-based methods (\textit{i.e.}, IRCNN~\cite{IRCNN}, FFDNet~\cite{ffdnet}, DnCNN~\cite{DnCNN} and DRUNet~\cite{drunet}), four transformer-based methods (\textit{i.e.}, SwinIR~\cite{SwinIR}, Restormer~\cite{Restormer}, CODE~\cite{CODE} and ART~\cite{art}) and one Mamba-based method (\textit{i.e.,} MambaIR~\cite{MambaIR}) as the baselines for synthetic noise removal. For real-world image denoising, four CNN-based methods (\textit{i.e.}, DeamNet~\cite{deamnet}, MPRNet~\cite{MPRNet}, NBNet~\cite{nbnet} and DAGL~\cite{dagl}), two transformer-based methods (\textit{i.e.}, Uformer~\cite{uformer} and Restormer~\cite{Restormer}) and one Mamba-based method (\textit{i.e.,} MambaIR~\cite{MambaIR}) are introduced for comparisons.

\textbf{Results:} As depicted in the~\cref{Tab:guassianDN}-\ref{Tab:realDN}, MaIR demonstrates superior performance on both synthetic and real-world image denoising compared to baselines. Taking results on Urban100 as examples, MaIR averagely outperforms MambaIR by 0.21dB in terms of PSNR, indicates its superiority on image denoising. Similar results can be derived from the qualitative comparisons shown in~\cref{Figure:denoise}, MaIR could keep more detailed textures on the restored images, which are more closely to the ground truth.


%

\begin{table}[t]
    \centering
    \normalsize
    \caption{Quantitative results on image motion deblurring. The best and second best results are in \textcolor{red}{red} and \textcolor{blue}{blue}. MACs in this table are evaluated on 128$\times$128 patches followed~\cite{CODE}.}
    \begin{tabular}{l | c | c | c | c }
	\toprule
         Method & Params & MACs & GoPro & HIDE \\
         \midrule
         SRN~\cite{srn} & 3.76M & 35.87G & 30.26 & 28.36 \\
         DBGAN~\cite{dbgan} & 11.59M & 379.92G & 31.10 & 28.94 \\
         MT-RNN~\cite{MT-RNN} & 2.64M & 13.72G & 31.15 & 29.15 \\
         DMPHN~\cite{dmphn} & 86.80M & - & 31.20 & 29.09 \\
         CODE~\cite{CODE} & 12.18M & 22.52G & 31.94 & 29.67 \\
         MIMO+~\cite{mimo} & 16.10M & 38.64G & 32.45 & 29.99 \\
         MPRNet~\cite{MPRNet} & 20.13M & 194.42G & 32.66 & 30.96 \\
         Restormer~\cite{Restormer} & 26.13M & 35.31G & 32.92 & 31.22 \\
         Uformer~\cite{uformer} & 50.88M & 22.36G & 33.06 & 30.90 \\
         CU-Mamba~\cite{cumamba} & 19.7M & - & 33.53 & \color{blue}{31.47}  \\
         NAFNet~\cite{nafnet} & 67.89M & 15.85G & \color{blue}{33.69} & 31.32 \\
         MaIR & 26.29M & 49.29G & \color{red}{33.69} & \color{red}{31.57} \\
         \bottomrule
    \end{tabular}
    \label{Tab:motiondeblur}
    \vspace{-0.5em}
\end{table}

\begin{figure*}[t]
\begin{center}
  \includegraphics[width=2.0\columnwidth]{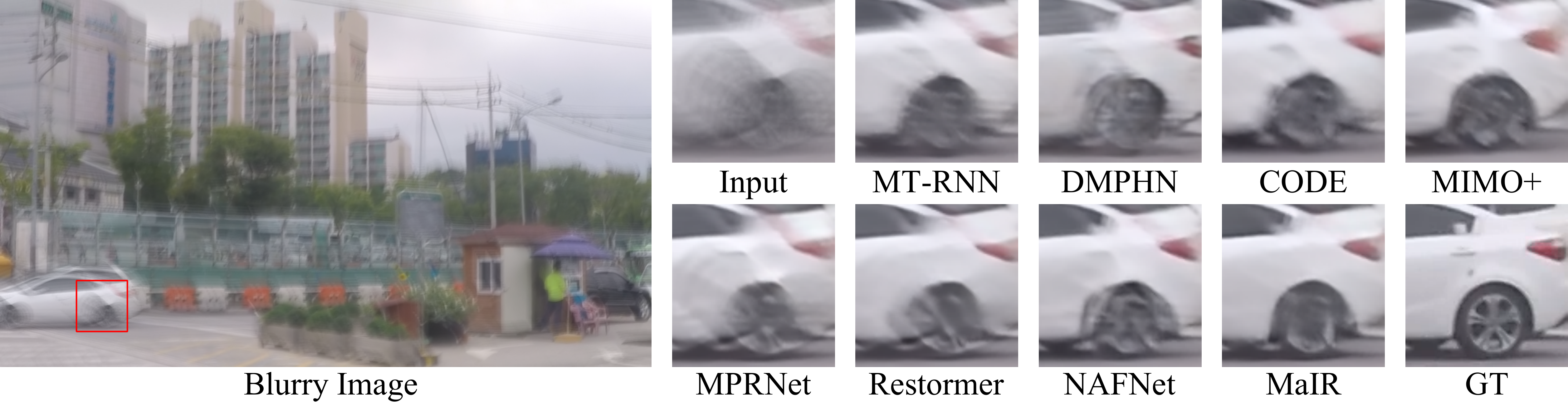}
\end{center}
\vspace{-1.5em}   
\caption{Visual comparison of motion deblurring results on the GoPro dataset. MaIR demonstrates superior performance in effectively removing motion blur while preserving precise fine details and textures, closely matching the ground truth.}
\label{Figure:deblur}
\end{figure*}

\subsection{Results on Image Deblurring}
In this section, we evaluate MaIR on motion deblurring to verify the effectiveness of our proposed method.

\textbf{Datasets:} Following previous works~\cite{MPRNet,Restormer}, we employ GoPro dataset~\cite{gopro} for training which consists of 2,103 blurry-clean image pairs. For evaluation, we use two common datasets, \textit{i.e.}, GoPro test set and HIDE~\cite{hide}, which consist of 1,111 and 2,025 blurry-clean pairs, respectively. 

\textbf{Baselines:} We adopt 11 competitive image deblurring baselines for comparisons. In detail, we adopt six CNN-based deblurring methods (\textit{i.e.}, SRN~\cite{srn}, DBGAN~\cite{dbgan}, DMPHN~\cite{dmphn}, MIMO~\cite{mimo}, MPRNet~\cite{MPRNet}, and NAFNet~\cite{nafnet}), three transformer-based methods (\textit{i.e.}, CODE~\cite{CODE}, Restormer~\cite{Restormer} and Uformer~\cite{uformer}), one RNN-based method (\textit{i.e.,} MT-RNN~\cite{MT-RNN}) and one Mamba-based method (\textit{i.e.}, CU-Mamba~\cite{cumamba}) as the baselines.

\textbf{Results:} As shown in~\cref{Tab:motiondeblur}, proposed MaIR surpasses other baselines by PSNR on both GoPro and HIDE. In detail, MaIR outperforms Restormer~\cite{Restormer} by 0.77dB on the GoPro dataset and by 0.35dB on the HIDE dataset in terms of PSNR. Although NAFNet achieves similar quantitative results on GoPro, MaIR surpasses NAFNet on HIDE dataset by 0.25dB in terms of PSNR. As illustrated in~\cref{Figure:deblur}, MaIR demonstrates its ability on handling heavily degraded areas, \textit{i.e.}, MaIR could effectively remove blur and restore details around the wheels.

\begin{table*}[t]
    \centering
    \normalsize
    \caption{Quantitative results on image dehazing. The best and second best results are in \textcolor{red}{red} and \textcolor{blue}{blue}. MACs in this table are evaluated on 256$\times$256 patches followed~\cite{dehazeformer,uvmnet}.
    }
 \resizebox{1.0\textwidth}{!}{
    \begin{tabular}{m{0.5cm} m{0.8cm}|c c c c c c c c c }
	\toprule
        \multicolumn{2}{c|}{\multirow{2}{*}{Method}} & {AODNet} & {GDN} & {MSBDN} & {FFANet} & AECRNet & Dehamer & {DehazeFormer}  & UVM-Net & {MaIR} \\
         & & {\cite{AOD-Net}} & {\cite{GridDehazeNet}} & {\cite{MSBDN}} & {\cite{FFA-Net}} & \cite{AECR-Net}  & \cite{dehamer} & {\cite{dehazeformer}} & \cite{uvmnet} & (Ours) \\
         \midrule
	   \multicolumn{2}{c|}{Params} & 0.002M & 0.96M & 31.35M & 4.46M & 2.61M & 132.45M & 4.63M & 1,003.94M & 3.40M \\
	   \multicolumn{2}{c|}{MACs}    
          & 0.115G & 21.49G & 41.54G & 287.8G & 52.20G & 48.93G & 48.64G & 501.91G & 24.03G \\
         \midrule
         \multicolumn{1}{c|}{SOTS-} & PSNR & 20.51  & 32.16  & 33.67  & 36.39 & 37.17 & 36.63 & 38.46 & \textcolor{red}{40.17} & \textcolor{blue}{39.45} \\
	      
         \multicolumn{1}{c|}{Indoor} & SSIM & 0.816 & 0.984 & 0.985 &  0.989 & 0.990 & 0.988 & 0.994 & \textcolor{blue}{0.996} & \textcolor{red}{0.997} \\
            \midrule

         \multicolumn{1}{c|}{SOTS-} & PSNR & 24.14 & 30.86 & 33.48 & 33.57 & - & 35.18 & 34.29 & \textcolor{blue}{34.92} & \textcolor{red}{36.96} \\
        \multicolumn{1}{c|}{Outdoor} & SSIM & 0.920 & 0.982 & 0.982 & 0.984 & - & 0.986 & 0.983 & \textcolor{blue}{0.984} & \textcolor{red}{0.991} \\
            \midrule

        \multicolumn{1}{c|}{SOTS-} & PSNR & 20.27 & 25.86  & 28.56 & 29.96 & 28.52 & - & 30.89 & \textcolor{red}{31.92} & \textcolor{blue}{31.52} \\
         \multicolumn{1}{c|}{Mix} & SSIM & 0.855 & 0.944 & 0.966 & 0.973 & 0.964  & - & 0.977 & \textcolor{red}{0.982} & \textcolor{blue}{0.980} \\
         \bottomrule
    \end{tabular}
    }
    \label{Tab:dehaze}
\end{table*}

\begin{figure*}[t]
\begin{center}
  \includegraphics[width=2.0\columnwidth]{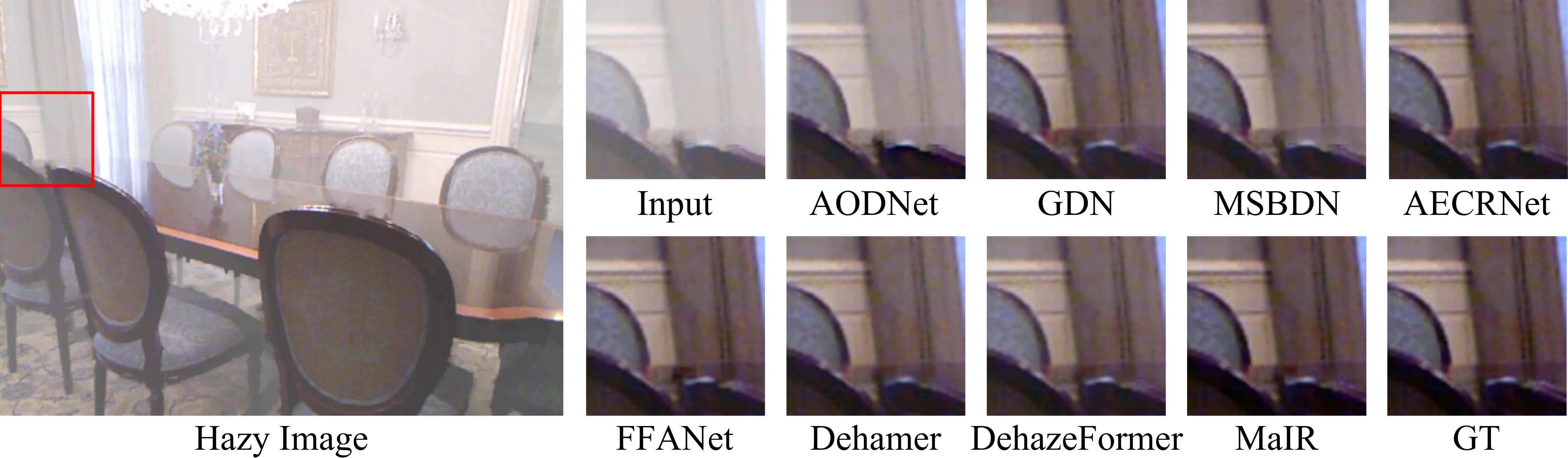}
\end{center}
\vspace{-1.5em}   
\caption{Visual comparison of image dehazing results on the SOTS dataset. MaIR can effectively remove haze and restore content with colors that closely match the ground truth. }
\vspace{-1.5em}   
\label{Figure:dehaze}
\end{figure*}

\subsection{Results on Image Dehazing}
In this section, we evaluate MaIR on image dehazing to verify the effectiveness of MaIR.

\textbf{Datasets:} Following existing works~\cite{dehazeformer}, we employ RESIDE dataset~\cite{RESIDE} for training and testing. For indoor scenes, we train MaIR on Indoor Training Set (ITS) which consists of 13,990 hazy-clean pairs, and test it on indoor synthetic objective testing set (SOTS-Indoor) involving 500 pairs. For outdoor scenes, we train MaIR on Outdoor Training Set (OTS), which contains 313,950 image pairs, and evaluate it on outdoor synthetic objective testing set (SOTS-Outdoor) involving 500 images. In addition, to verify MaIR on more general cases, we also train the model on RESIDE-6K and test it on the SOTS-mix, which mix both indoor and outdoor images. 

\textbf{Baselines:} We adopt eight competitive methods as baselines. Specifically, we adopt five CNN-based image dehazing methods (\textit{i.e.}, AODNet~\cite{AOD-Net}, GDN~\cite{GridDehazeNet}, MSBDN~\cite{MSBDN}, FFANet~\cite{FFA-Net} and AECRNet~\cite{AECR-Net}), two transformer-based methods (\textit{i.e.}, Dehamer~\cite{dehamer}, Dehazeformer~\cite{dehazeformer}) and one Mamba-based method (\textit{i.e.,} UVM-Net~\cite{uvmnet}) as baselines.

\textbf{Results:} As shown in \cref{Tab:dehaze,Figure:dehaze}, our MaIR surpasses most baselines on both quantitative and qualitative comparisons. Taking quantitative results as examples, MaIR significantly outperforms DehazeFormer and UVM-Net by 2.67dB and 2.04dB in terms of PSNR on the outdoor scenes. Although UVM-Net is slightly higher on PSNR in the indoor scenes, MaIR only takes 0.3\% and 4.8\% params and MACs of the UVM-Net, which verifies both effectiveness and efficiency.

\subsection{Analysis Experiments}
In this section, we first conduct ablation studies to verify the effectiveness of the NSS and SSA. Then, we introduce analysis experiments to verify the observations and investigate the impact of stripe width on the overall performance.


\begin{table}[t!]
    \centering
    \footnotesize
    \caption{Ablation study on NSS, tested on lightweight super-resolution with scale factor $\times2$. The results on the Urban100 dataset are presented, which demonstrates the effectiveness of the NSS.}
    \vspace{-0.5em}
    \begin{tabular}{w{c}{0.4cm} | c c c c c c}
	\toprule
          & Baseline & w/o NSS & w/o SS & LM & ZigMa & PH\\
         \midrule
         PSNR & 32.97 & 32.94 & 32.93 & 32.93 & 32.88 & 32.95\\
         SSIM & 0.9359 & 0.9355  & 0.9351 & 0.9357 & 0.9354 & 0.9356\\
         \bottomrule
    \end{tabular}
    \label{Tab:ab_nsss}
\end{table}

\subsubsection{Ablation Studies}
We first conduct ablation study to analyze the effectiveness of NSS. In detail, five configurations are conducted, i) replacing NSS with Z-shaped scanning strategy (denoted as w/o NSS), ii) removing shift stripe (denoted as w/o SS), iii) replacing NSS with the scanning strategy in LocalMamba~\cite{localmamba} (denoted as LM), iv) replacing NSS by the scanning strategy in ZigMa~\cite{zigma} (denoted as ZigMa) and v) replacing NSS by the Peano-Hilbert curve (denoted as PH). As illustrated in~\cref{Tab:ab_nsss}, NSS is important to improve the performance of MaIR.


\begin{table}[t!]
    \centering
    \normalsize
    \caption{Ablation study on SSA for lightweight super-resolution with scale factor $\times2$. The results on Urban100 dataset demonstrate the effectiveness of SSA.}
    \vspace{-0.5em}
    \begin{tabular}{l | c c c c}
	\toprule
          & MaIR & w/o SSA & UVM & SeqGat \\
         \midrule
         PSNR & 32.97 & 32.90 & 32.59 & 32.92 \\
         SSIM & 0.9359 & 0.9350 & 0.9324 & 0.9352 \\
         \midrule
          & CAGat & FPixGat & DWPixGat & -  \\
         \midrule
         PSNR & 32.71 & 31.90 & 32.74 & - \\
         SSIM & 0.9335 & 0.9250 &  0.9342 & - \\
         \bottomrule
    \end{tabular}
    \label{Tab:ab_SSA}
    \vspace{-1em}
\end{table}

To investigate the effectiveness of SSA, we remove SSA and aggregate sequences through: i) sequences-wise addition (termed as w/o SSA), ii) SSM~\cite{uvmnet} (termed as UVM), iii) sequence-wise gating~\cite{rsmamba} (termed as SeqGat), 
iv) channel-wise gating (termed as CAGat), 
v) pixel-wise gating through fully connected convolution (termed as FPixGat). 
vi) pixel-wise gating through depth-wise convolution (termed as DWPixGat). 
It is worth noting that we keep the size of different models to be similar for fair comparisons.
As shown in~\cref{Tab:ab_SSA}, SSA is more effective than others.


\begin{table}[t!]
    \centering
    \normalsize
    \caption{Analyses on stripe widths. Experiment was conducted on the Urban100 dataset with a scale factor $\times2$ for lightweight super-resolution tasks, which illustrates how changes in stripe width affect the restored image quality.}
    \vspace{-0.5em}
    \begin{tabular}{l | c  c  c c c}
	\toprule
          & 2 & 4 & 8 & 16 & 32\\
         \midrule
         PSNR & 32.95 & 32.97 & 32.97 & 32.97 & 32.92\\
         SSIM & 0.9356 & 0.9359 & 0.9355 & 0.9357 & 0.9355\\
         \bottomrule
    \end{tabular}
    \label{Tab:scan_width}
\end{table}

\begin{figure*}[t]
\begin{center}
  \includegraphics[width=1.0\linewidth]{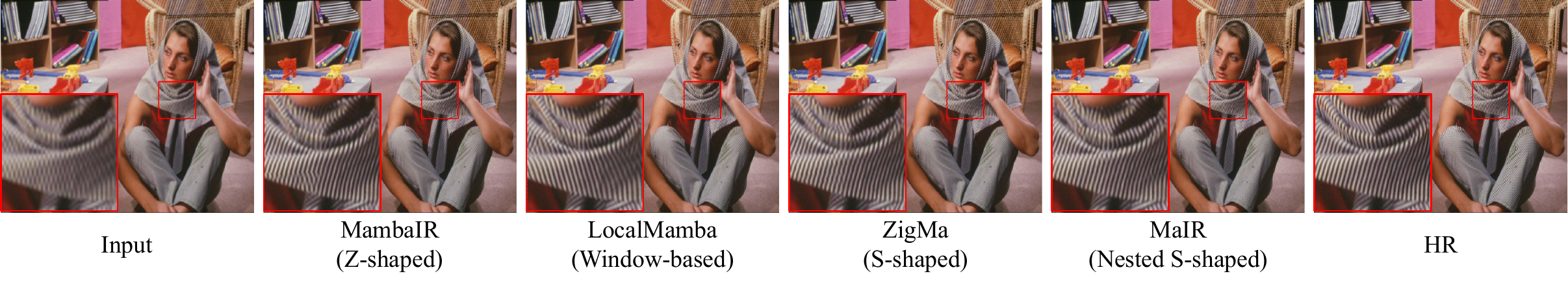}
\end{center}
\vspace{-2em}   
\caption{Visual comparisons of different scanning strategies, illustrating that i) windows-based scanning path overlooks the continuity between different regions (\eg, relationship between different layers of the scarf), resulting in wrong textures, ii) S-shaped scanning path leads to distortion in local regions, causing the scarf’s texture to appear warped. iii) Z-shaped scanning path suffers from both of them. In contrast, MaIR avoids aforementioned problems and achieves visually appealing results.}
\vspace{-0.5em}   
\label{Figure:vis_ob}
\end{figure*}

\subsubsection{Verification of Observations}
To verify observations shown in~\cref{Figure:observation}, we conduct visual comparisons among different scanning strategies. As shown in~\cref{Figure:vis_ob}, proposed methods can maintain both locality and continuity and produce more visual pleasant results.

\subsubsection{Results on different stripe width}
To investigate influence of stripe width, we train lightweight SR model with stripe width $w_s=\{2, 4, 8, 16, 32\}$ and evaluate them on Urban100 dataset. As presented in \cref{Tab:scan_width}, the PSNR and SSIM values are quite similar under different settings, except for the cases with the largest and the smallest stripe widths. It indicates that the proposed method exhibits robustness against changes in stripe width, maintaining high-quality image restoration across a range of stripe widths.





\section{Conclusion}
In this paper, we propose MaIR, a novel state space model for image restoration that can preserve both local dependencies and spatial continuity of input images. To this end, we propose two designs: Nested S-shaped Scanning strategy (NSS) and Sequences Shuffle Attention (SSA). NSS is designed to extract locality- and continuity-preserving sequences from images, and SSA adaptively aggregates these sequences. Thanks to their cooperation, MaIR not only addresses the limitations of existing Mamba-based restoration methods but also improves image quality without introducing extra computations. Extensive experiments across four tasks on 14 benchmarks comparing with 40 baselines validate the superiority of MaIR, demonstrating its robustness and effectiveness in various image restoration tasks. 

\section*{Acknowledgments}
This work was supported in part by NSFC under Grant 62176171, U21B2040, 62472295; in part by the Fundamental Research Funds for the Central Universities under Grant CJ202303; and in part by Sichuan Science and Technology Planning Project under Grant 24NSFTD0130.

{
    \small
    \bibliographystyle{ieeenat_fullname}
    \bibliography{main}

\begin{thebibliography}{70}
\providecommand{\natexlab}[1]{#1}
\providecommand{\url}[1]{\texttt{#1}}
\expandafter\ifx\csname urlstyle\endcsname\relax
  \providecommand{\doi}[1]{doi: #1}\else
  \providecommand{\doi}{doi: \begingroup \urlstyle{rm}\Url}\fi

\bibitem[Abdelhamed et~al.(2018)Abdelhamed, Lin, and Brown]{sidd}
Abdelrahman Abdelhamed, Stephen Lin, and Michael~S Brown.
\newblock A high-quality denoising dataset for smartphone cameras.
\newblock In \emph{Proceedings of the IEEE conference on computer vision and pattern recognition}, pages 1692--1700, 2018.

\bibitem[Ahn et~al.(2018)Ahn, Kang, and Sohn]{CARN}
Namhyuk Ahn, Byungkon Kang, and Kyung-Ah Sohn.
\newblock Fast, accurate, and lightweight super-resolution with cascading residual network.
\newblock In \emph{Proceedings of the European conference on computer vision (ECCV)}, pages 252--268, 2018.

\bibitem[Bevilacqua et~al.(2012)Bevilacqua, Roumy, Guillemot, and Alberi-Morel]{set5}
Marco Bevilacqua, Aline Roumy, Christine Guillemot, and Marie~Line Alberi-Morel.
\newblock Low-complexity single-image super-resolution based on nonnegative neighbor embedding.
\newblock 2012.

\bibitem[Chen et~al.(2021)Chen, Wang, Guo, Xu, Deng, Liu, Ma, Xu, Xu, and Gao]{IPT}
Hanting Chen, Yunhe Wang, Tianyu Guo, Chang Xu, Yiping Deng, Zhenhua Liu, Siwei Ma, Chunjing Xu, Chao Xu, and Wen Gao.
\newblock {Pre-Trained Image Processing Transformer}.
\newblock In \emph{IEEE Conference on Computer Vision and Pattern Recognition}, pages 12299--12310, Virtual, 2021.

\bibitem[Chen et~al.(2024)Chen, Chen, Liu, Li, Zou, and Shi]{rsmamba}
Keyan Chen, Bowen Chen, Chenyang Liu, Wenyuan Li, Zhengxia Zou, and Zhenwei Shi.
\newblock Rsmamba: Remote sensing image classification with state space model.
\newblock \emph{IEEE Geoscience and Remote Sensing Letters}, 2024.

\bibitem[Chen et~al.(2022)Chen, Chu, Zhang, and Sun]{nafnet}
Liangyu Chen, Xiaojie Chu, Xiangyu Zhang, and Jian Sun.
\newblock Simple baselines for image restoration.
\newblock In \emph{European conference on computer vision}, pages 17--33. Springer, 2022.

\bibitem[Chen et~al.(2023)Chen, Wang, Zhou, Qiao, and Dong]{hat}
Xiangyu Chen, Xintao Wang, Jiantao Zhou, Yu Qiao, and Chao Dong.
\newblock Activating more pixels in image super-resolution transformer.
\newblock In \emph{Proceedings of the IEEE/CVF conference on computer vision and pattern recognition}, pages 22367--22377, 2023.

\bibitem[Cheng et~al.(2021)Cheng, Wang, Huang, Liu, Fan, and Liu]{nbnet}
Shen Cheng, Yuzhi Wang, Haibin Huang, Donghao Liu, Haoqiang Fan, and Shuaicheng Liu.
\newblock Nbnet: Noise basis learning for image denoising with subspace projection.
\newblock In \emph{Proceedings of the IEEE/CVF conference on computer vision and pattern recognition}, pages 4896--4906, 2021.

\bibitem[Cho et~al.(2021)Cho, Ji, Hong, Jung, and Ko]{mimo}
Sung-Jin Cho, Seo-Won Ji, Jun-Pyo Hong, Seung-Won Jung, and Sung-Jea Ko.
\newblock Rethinking coarse-to-fine approach in single image deblurring.
\newblock In \emph{Proceedings of the IEEE/CVF international conference on computer vision}, pages 4641--4650, 2021.

\bibitem[Dai et~al.(2019)Dai, Cai, Zhang, Xia, and Zhang]{SAN}
Tao Dai, Jianrui Cai, Yongbing Zhang, Shu-Tao Xia, and Lei Zhang.
\newblock Second-order attention network for single image super-resolution.
\newblock In \emph{IEEE Conference on Computer Vision and Pattern Recognition}, pages 11065--11074, 2019.

\bibitem[Dao and Gu(2024)]{mamba2}
Tri Dao and Albert Gu.
\newblock Transformers are {SSM}s: Generalized models and efficient algorithms through structured state space duality.
\newblock In \emph{International Conference on Machine Learning (ICML)}, 2024.

\bibitem[Deng and Gu(2024)]{cumamba}
Rui Deng and Tianpei Gu.
\newblock Cu-mamba: Selective state space models with channel learning for image restoration.
\newblock \emph{arXiv preprint arXiv:2404.11778}, 2024.

\bibitem[Dong et~al.(2020)Dong, Pan, Xiang, Hu, Zhang, Wang, and Yang]{MSBDN}
Hang Dong, Jinshan Pan, Lei Xiang, Zhe Hu, Xinyi Zhang, Fei Wang, and Ming-Hsuan Yang.
\newblock {Multi-Scale Boosted Dehazing Network with Dense Feature Fusion}.
\newblock In \emph{IEEE Conference on Computer Vision and Pattern Recognition}, pages 2154--2164, Seattle, WA, 2020.

\bibitem[Gou et~al.(2020)Gou, Li, Liu, Yang, and Peng]{clear}
Yuanbiao Gou, Boyun Li, Zitao Liu, Songfan Yang, and Xi Peng.
\newblock Clearer: Multi-scale neural architecture search for image restoration.
\newblock \emph{Advances in Neural Information Processing Systems}, 33, 2020.

\bibitem[Gu and Dao(2023)]{mamba}
Albert Gu and Tri Dao.
\newblock Mamba: Linear-time sequence modeling with selective state spaces.
\newblock \emph{arXiv preprint arXiv:2312.00752}, 2023.

\bibitem[Gu et~al.(2021)Gu, Goel, and R{\'e}]{SSM}
Albert Gu, Karan Goel, and Christopher R{\'e}.
\newblock Efficiently modeling long sequences with structured state spaces.
\newblock \emph{arXiv preprint arXiv:2111.00396}, 2021.

\bibitem[Guo et~al.(2022)Guo, Yan, Anwar, Cong, Ren, and Chongyi]{dehamer}
Chun-Le Guo, Qixin Yan, Saeed Anwar, Runmin Cong, Wenqi Ren, and Li Chongyi.
\newblock Image dehazing transformer with transmission-aware 3d position embedding.
\newblock In \emph{Proceedings of the IEEE/CVF Conference on Computer Vision and Pattern Recognition}, 2022.

\bibitem[Guo et~al.(2024)Guo, Li, Dai, Ouyang, Ren, and Xia]{MambaIR}
Hang Guo, Jinmin Li, Tao Dai, Zhihao Ouyang, Xudong Ren, and Shu-Tao Xia.
\newblock Mambair: A simple baseline for image restoration with state-space model.
\newblock \emph{arXiv preprint arXiv:2402.15648}, 2024.

\bibitem[Hu et~al.(2024)Hu, Baumann, Gui, Grebenkova, Ma, Fischer, and Ommer]{zigma}
Vincent~Tao Hu, Stefan~Andreas Baumann, Ming Gui, Olga Grebenkova, Pingchuan Ma, Johannes~S Fischer, and Bj{\"o}rn Ommer.
\newblock Zigma: A dit-style zigzag mamba diffusion model.
\newblock \emph{arXiv preprint arXiv:2403.13802}, 2024.

\bibitem[Huang et~al.(2015)Huang, Singh, and Ahuja]{urban100}
Jia-Bin Huang, Abhishek Singh, and Narendra Ahuja.
\newblock Single image super-resolution from transformed self-exemplars.
\newblock In \emph{Proceedings of the IEEE conference on computer vision and pattern recognition}, pages 5197--5206, 2015.

\bibitem[Huang et~al.(2024)Huang, Pei, You, Wang, Qian, and Xu]{localmamba}
Tao Huang, Xiaohuan Pei, Shan You, Fei Wang, Chen Qian, and Chang Xu.
\newblock Localmamba: Visual state space model with windowed selective scan.
\newblock \emph{arXiv preprint arXiv:2403.09338}, 2024.

\bibitem[Hui et~al.(2019)Hui, Gao, Yang, and Wang]{IMDN}
Zheng Hui, Xinbo Gao, Yunchu Yang, and Xiumei Wang.
\newblock Lightweight image super-resolution with information multi-distillation network.
\newblock In \emph{Proceedings of the 27th acm international conference on multimedia}, pages 2024--2032, 2019.

\bibitem[Li et~al.(2017)Li, Peng, Wang, Xu, and Feng]{AOD-Net}
Boyi Li, Xiulian Peng, Zhangyang Wang, Jizheng Xu, and Dan Feng.
\newblock {AOD-Net: All-in-One Dehazing Network}.
\newblock In \emph{IEEE International Conference on Computer Vision}, pages 4780--4788, Venice, Italy, 2017.

\bibitem[Li et~al.(2019)Li, Ren, Fu, Tao, Feng, Zeng, and Wang]{RESIDE}
Boyi Li, Wenqi Ren, Dengpan Fu, Dacheng Tao, Dan Feng, Wenjun Zeng, and Zhangyang Wang.
\newblock {Benchmarking Single Image Dehazing and Beyond}.
\newblock \emph{IEEE Transactions on Image Processing}, 28\penalty0 (1):\penalty0 492--505, 2019.

\bibitem[Li et~al.(2022)Li, Liu, Hu, Wu, Lv, and Peng]{AirNet}
Boyun Li, Xiao Liu, Peng Hu, Zhongqin Wu, Jiancheng Lv, and Xi Peng.
\newblock {All-In-One Image Restoration for Unknown Corruption}.
\newblock In \emph{IEEE Conference on Computer Vision and Pattern Recognition}, pages 17431--17441, New Orleans, LA, 2022.

\bibitem[Li et~al.(2020)Li, Zhou, Qi, Jiang, Lu, and Jia]{LAPAR}
Wenbo Li, Kun Zhou, Lu Qi, Nianjuan Jiang, Jiangbo Lu, and Jiaya Jia.
\newblock Lapar: Linearly-assembled pixel-adaptive regression network for single image super-resolution and beyond.
\newblock \emph{Advances in Neural Information Processing Systems}, 33:\penalty0 20343--20355, 2020.

\bibitem[Liang et~al.(2021)Liang, Cao, Sun, Zhang, Van~Gool, and Timofte]{SwinIR}
Jingyun Liang, Jiezhang Cao, Guolei Sun, Kai Zhang, Luc Van~Gool, and Radu Timofte.
\newblock {SwinIR: Image Restoration Using Swin Transformer}.
\newblock In \emph{International Conference on Computer Vision Workshops}, Virtual, 2021.

\bibitem[Lim et~al.(2017{\natexlab{a}})Lim, Son, Kim, Nah, and Lee]{EDSR}
Bee Lim, Sanghyun Son, Heewon Kim, Seungjun Nah, and Kyoung~Mu Lee.
\newblock Enhanced deep residual networks for single image super-resolution.
\newblock In \emph{IEEE Conference on Computer Vision and Pattern Recognition WorkShop}, pages 1132--1140, 2017{\natexlab{a}}.

\bibitem[Lim et~al.(2017{\natexlab{b}})Lim, Son, Kim, Nah, and Mu~Lee]{flickr2k}
Bee Lim, Sanghyun Son, Heewon Kim, Seungjun Nah, and Kyoung Mu~Lee.
\newblock Enhanced deep residual networks for single image super-resolution.
\newblock In \emph{Proceedings of the IEEE conference on computer vision and pattern recognition workshops}, pages 136--144, 2017{\natexlab{b}}.

\bibitem[Liu et~al.(2019)Liu, Ma, Shi, and Chen]{GridDehazeNet}
Xiaohong Liu, Yongrui Ma, Zhihao Shi, and Jun Chen.
\newblock {GridDehazeNet: Attention-Based Multi-Scale Network for Image Dehazing}.
\newblock In \emph{International Conference on Computer Vision}, pages 7313--7322, Seoul, Korea, 2019.

\bibitem[Liu et~al.(2024)Liu, Tian, Zhao, Yu, Xie, Wang, Ye, and Liu]{vmamba}
Yue Liu, Yunjie Tian, Yuzhong Zhao, Hongtian Yu, Lingxi Xie, Yaowei Wang, Qixiang Ye, and Yunfan Liu.
\newblock Vmamba: Visual state space model.
\newblock \emph{arXiv preprint arXiv:2401.10166}, 2024.

\bibitem[Liu et~al.(2021)Liu, Lin, Cao, Hu, Wei, Zhang, Lin, and Guo]{swin}
Ze Liu, Yutong Lin, Yue Cao, Han Hu, Yixuan Wei, Zheng Zhang, Stephen Lin, and Baining Guo.
\newblock Swin transformer: Hierarchical vision transformer using shifted windows.
\newblock In \emph{Proceedings of the IEEE/CVF international conference on computer vision}, pages 10012--10022, 2021.

\bibitem[Luo et~al.(2020)Luo, Xie, Zhang, Qu, Li, and Fu]{Latticenet}
Xiaotong Luo, Yuan Xie, Yulun Zhang, Yanyun Qu, Cuihua Li, and Yun Fu.
\newblock Latticenet: Towards lightweight image super-resolution with lattice block.
\newblock In \emph{European Conference on Computer Vision}, pages 272--289, 2020.

\bibitem[Ma et~al.(2017)Ma, Duanmu, Wu, Wang, Yong, Li, and Zhang]{WED}
Kede Ma, Zhengfang Duanmu, Qingbo Wu, Zhou Wang, Hongwei Yong, Hongliang Li, and Lei Zhang.
\newblock {Waterloo Exploration Database: New Challenges for Image Quality Assessment Models}.
\newblock \emph{IEEE Transactions on Image Processing}, 26\penalty0 (2):\penalty0 1004--1016, 2017.

\bibitem[Martin et~al.(2001)Martin, Fowlkes, Tal, and Malik]{BSD}
David Martin, Charless Fowlkes, Doron Tal, and Jitendra Malik.
\newblock {A Database of Human Segmented Natural Images and its Application to Evaluating Segmentation Algorithms and Measuring Ecological Statistics}.
\newblock In \emph{International Conference on Computer Vision}, pages 416--425, Vancouver, Canada, 2001.

\bibitem[Matsui et~al.(2017)Matsui, Ito, Aramaki, Fujimoto, Ogawa, Yamasaki, and Aizawa]{manga109}
Yusuke Matsui, Kota Ito, Yuji Aramaki, Azuma Fujimoto, Toru Ogawa, Toshihiko Yamasaki, and Kiyoharu Aizawa.
\newblock Sketch-based manga retrieval using manga109 dataset.
\newblock \emph{Multimedia tools and applications}, 76:\penalty0 21811--21838, 2017.

\bibitem[Mei et~al.(2021)Mei, Fan, and Zhou]{NLSA}
Yiqun Mei, Yuchen Fan, and Yuqian Zhou.
\newblock Image super-resolution with non-local sparse attention.
\newblock In \emph{IEEE Conference on Computer Vision and Pattern Recognition}, pages 3517--3526, 2021.

\bibitem[Mou et~al.(2021)Mou, Zhang, and Wu]{dagl}
Chong Mou, Jian Zhang, and Zhuoyuan Wu.
\newblock Dynamic attentive graph learning for image restoration.
\newblock In \emph{IEEE International Conference on Computer Vision}, 2021.

\bibitem[Nah et~al.(2017)Nah, Kim, and Lee]{gopro}
Seungjun Nah, Tae~Hyun Kim, and Kyoung~Mu Lee.
\newblock {Deep Multi-scale Convolutional Neural Network for Dynamic Scene Deblurring}.
\newblock In \emph{IEEE Conference on Computer Vision and Pattern Recognition}, pages 257--265, Honolulu, HI, 2017.

\bibitem[Niu et~al.(2020)Niu, Wen, Ren, Zhang, Yang, Wang, Zhang, Cao, and Shen]{HAN}
Ben Niu, Weiwei Wen, Wenqi Ren, Xiangde Zhang, Lianping Yang, Shuzhen Wang, Kaihao Zhang, Xiaochun Cao, and Haifeng Shen.
\newblock Single image super-resolution via a holistic attention network.
\newblock In \emph{European Conference on Computer Vision}, pages 191--207, 2020.

\bibitem[Park et~al.(2020)Park, Kang, Kim, and Chun]{MT-RNN}
Dongwon Park, Dong~Un Kang, Jisoo Kim, and Se~Young Chun.
\newblock Multi-temporal recurrent neural networks for progressive non-uniform single image deblurring with incremental temporal training.
\newblock In \emph{European Conference on Computer Vision}, pages 327--343. Springer, 2020.

\bibitem[Qin et~al.(2020)Qin, Wang, Bai, Xie, and Jia]{FFA-Net}
Xu Qin, Zhilin Wang, Yuanchao Bai, Xiaodong Xie, and Huizhu Jia.
\newblock {FFA-Net: Feature Fusion Attention Network for Single Image Dehazing}.
\newblock In \emph{AAAI Conference on Artificial Intelligence}, pages 11908--11915, New York, NY, 2020.

\bibitem[Ren et~al.(2021)Ren, He, Wang, and Zhao]{deamnet}
Chao Ren, Xiaohai He, Chuncheng Wang, and Zhibo Zhao.
\newblock Adaptive consistency prior based deep network for image denoising.
\newblock In \emph{Proceedings of the IEEE/CVF conference on computer vision and pattern recognition}, pages 8596--8606, 2021.

\bibitem[Sakaridis et~al.(2018)Sakaridis, Dai, and Van~Gool]{FDD}
Christos Sakaridis, Dengxin Dai, and Luc Van~Gool.
\newblock Semantic foggy scene understanding with synthetic data.
\newblock \emph{International Journal of Computer Vision}, 126\penalty0 (9):\penalty0 973--992, 2018.

\bibitem[Shen et~al.(2019)Shen, Wang, Lu, Shen, Ling, Xu, and Shao]{hide}
Ziyi Shen, Wenguan Wang, Xiankai Lu, Jianbing Shen, Haibin Ling, Tingfa Xu, and Ling Shao.
\newblock Human-aware motion deblurring.
\newblock In \emph{Proceedings of the IEEE/CVF international conference on computer vision}, pages 5572--5581, 2019.

\bibitem[Smith et~al.(2022)Smith, Warrington, and Linderman]{s5}
Jimmy~TH Smith, Andrew Warrington, and Scott Linderman.
\newblock Simplified state space layers for sequence modeling.
\newblock In \emph{The Eleventh International Conference on Learning Representations}, 2022.

\bibitem[Song et~al.(2023)Song, He, Qian, and Du]{dehazeformer}
Yuda Song, Zhuqing He, Hui Qian, and Xin Du.
\newblock Vision transformers for single image dehazing.
\newblock \emph{IEEE Transactions on Image Processing}, 32:\penalty0 1927--1941, 2023.

\bibitem[Tao et~al.(2018)Tao, Gao, Shen, Wang, and Jia]{srn}
Xin Tao, Hongyun Gao, Xiaoyong Shen, Jue Wang, and Jiaya Jia.
\newblock Scale-recurrent network for deep image deblurring.
\newblock In \emph{Proceedings of the IEEE conference on computer vision and pattern recognition}, pages 8174--8182, 2018.

\bibitem[Timofte et~al.(2017)Timofte, Agustsson, Van~Gool, Yang, and Zhang]{div2k}
Radu Timofte, Eirikur Agustsson, Luc Van~Gool, Ming-Hsuan Yang, and Lei Zhang.
\newblock Ntire 2017 challenge on single image super-resolution: Methods and results.
\newblock In \emph{Proceedings of the IEEE conference on computer vision and pattern recognition workshops}, pages 114--125, 2017.

\bibitem[Wang et~al.(2022)Wang, Cun, Bao, Zhou, Liu, and Li]{uformer}
Zhendong Wang, Xiaodong Cun, Jianmin Bao, Wengang Zhou, Jianzhuang Liu, and Houqiang Li.
\newblock Uformer: A general u-shaped transformer for image restoration.
\newblock In \emph{Proceedings of the IEEE/CVF Conference on Computer Vision and Pattern Recognition (CVPR)}, pages 17683--17693, 2022.

\bibitem[Wu et~al.(2021)Wu, Qu, Lin, Zhou, Qiao, Zhang, Xie, and Ma]{AECR-Net}
Haiyan Wu, Yanyun Qu, Shaohui Lin, Jian Zhou, Ruizhi Qiao, Zhizhong Zhang, Yuan Xie, and Lizhuang Ma.
\newblock {Contrastive Learning for Compact Single Image Dehazing}.
\newblock In \emph{IEEE Conference on Computer Vision and Pattern Recognition}, pages 10551--10560, Virtual, 2021.

\bibitem[Zamir et~al.(2021)Zamir, Arora, Khan, Hayat, Khan, Yang, and Shao]{MPRNet}
Syed~Waqas Zamir, Aditya Arora, Salman Khan, Munawar Hayat, Fahad~Shahbaz Khan, Ming-Hsuan Yang, and Ling Shao.
\newblock {Multi-Stage Progressive Image Restoration}.
\newblock In \emph{IEEE Conference on Computer Vision and Pattern Recognition}, pages 14821--14831, Virtual, 2021.

\bibitem[Zamir et~al.(2022)Zamir, Arora, Khan, Hayat, Khan, and Yang]{Restormer}
Syed~Waqas Zamir, Aditya Arora, Salman Khan, Munawar Hayat, Fahad~Shahbaz Khan, and Ming-Hsuan Yang.
\newblock {Restormer: Efficient Transformer for High-Resolution Image Restoration}.
\newblock In \emph{IEEE Conference on Computer Vision and Pattern Recognition}, pages 5718--5729, New Orleans, LA, 2022.

\bibitem[Zeyde et~al.(2012)Zeyde, Elad, and Protter]{set14}
Roman Zeyde, Michael Elad, and Matan Protter.
\newblock On single image scale-up using sparse-representations.
\newblock In \emph{Curves and Surfaces: 7th International Conference, Avignon, France, June 24-30, 2010, Revised Selected Papers 7}, pages 711--730. Springer, 2012.

\bibitem[Zhang et~al.(2019)Zhang, Dai, Li, and Koniusz]{dmphn}
Hongguang Zhang, Yuchao Dai, Hongdong Li, and Piotr Koniusz.
\newblock Deep stacked hierarchical multi-patch network for image deblurring.
\newblock In \emph{Proceedings of the IEEE/CVF conference on computer vision and pattern recognition}, pages 5978--5986, 2019.

\bibitem[Zhang et~al.(2023)Zhang, Zhang, Gu, Zhang, Kong, and Yuan]{art}
Jiale Zhang, Yulun Zhang, Jinjin Gu, Yongbing Zhang, Linghe Kong, and Xin Yuan.
\newblock Accurate image restoration with attention retractable transformer.
\newblock In \emph{ICLR}, 2023.

\bibitem[Zhang et~al.(2017{\natexlab{a}})Zhang, Zuo, Chen, Meng, and Zhang]{DnCNN}
Kai Zhang, Wangmeng Zuo, Yunjin Chen, Deyu Meng, and Lei Zhang.
\newblock {Beyond a Gaussian Denoiser: Residual Learning of Deep CNN for Image Denoising}.
\newblock \emph{IEEE Transactions on Image Processing}, 26\penalty0 (7):\penalty0 3142--3155, 2017{\natexlab{a}}.

\bibitem[Zhang et~al.(2017{\natexlab{b}})Zhang, Zuo, Gu, and Zhang]{IRCNN}
Kai Zhang, Wangmeng Zuo, Shuhang Gu, and Lei Zhang.
\newblock Learning deep cnn denoiser prior for image restoration.
\newblock In \emph{IEEE Conference on Computer Vision and Pattern Recognition}, pages 3929--3938, 2017{\natexlab{b}}.

\bibitem[Zhang et~al.(2018{\natexlab{a}})Zhang, Zuo, and Zhang]{ffdnet}
Kai Zhang, Wangmeng Zuo, and Lei Zhang.
\newblock Ffdnet: Toward a fast and flexible solution for {CNN} based image denoising.
\newblock \emph{IEEE Transactions on Image Processing}, 2018{\natexlab{a}}.

\bibitem[Zhang et~al.(2020)Zhang, Luo, Zhong, Ma, Stenger, Liu, and Li]{dbgan}
Kaihao Zhang, Wenhan Luo, Yiran Zhong, Lin Ma, Bjorn Stenger, Wei Liu, and Hongdong Li.
\newblock Deblurring by realistic blurring.
\newblock In \emph{Proceedings of the IEEE/CVF conference on computer vision and pattern recognition}, pages 2737--2746, 2020.

\bibitem[Zhang et~al.(2021)Zhang, Li, Zuo, Zhang, Van~Gool, and Timofte]{drunet}
Kai Zhang, Yawei Li, Wangmeng Zuo, Lei Zhang, Luc Van~Gool, and Radu Timofte.
\newblock Plug-and-play image restoration with deep denoiser prior.
\newblock \emph{IEEE Transactions on Pattern Analysis and Machine Intelligence}, 44\penalty0 (10):\penalty0 6360--6376, 2021.

\bibitem[Zhang et~al.(2011)Zhang, Wu, Buades, and Li]{mcmaster}
Lei Zhang, Xiaolin Wu, Antoni Buades, and Xin Li.
\newblock Color demosaicking by local directional interpolation and nonlocal adaptive thresholding.
\newblock \emph{Journal of Electronic imaging}, 20\penalty0 (2):\penalty0 023016--023016, 2011.

\bibitem[Zhang et~al.(2022)Zhang, Zeng, Guo, and Zhang]{ELAN}
Xindong Zhang, Hui Zeng, Shi Guo, and Lei Zhang.
\newblock Efficient long-range attention network for image super-resolution.
\newblock In \emph{European Conference on Computer Vision}, pages 649--667. Springer, 2022.

\bibitem[Zhang et~al.(2018{\natexlab{b}})Zhang, Li, Li, Wang, Zhong, and Fu]{RCAN}
Yulun Zhang, Kunpeng Li, Kai Li, Lichen Wang, Bineng Zhong, and Yun Fu.
\newblock Image super-resolution using very deep residual channel attention networks.
\newblock In \emph{European Conference on Computer Vision}, pages 294--310, 2018{\natexlab{b}}.
\newblock To appear in ECCV 2018.

\bibitem[Zhang et~al.(2018{\natexlab{c}})Zhang, Tian, Kong, Zhong, and Fu]{RDN}
Yulun Zhang, Yapeng Tian, Yu Kong, Bineng Zhong, and Yun Fu.
\newblock {Residual Dense Network for Image Restoration}.
\newblock \emph{IEEE Transactions on Pattern Analysis and Machine Intelligence}, 43\penalty0 (7):\penalty0 2480--2495, 2018{\natexlab{c}}.

\bibitem[Zhao et~al.(2023)Zhao, Gou, Li, Peng, Lv, and Peng]{CODE}
Haiyu Zhao, Yuanbiao Gou, Boyun Li, Dezhong Peng, Jiancheng Lv, and Xi Peng.
\newblock Comprehensive and delicate: An efficient transformer for image restoration.
\newblock In \emph{Proceedings of the IEEE/CVF Conference on Computer Vision and Pattern Recognition}, pages 14122--14132, 2023.

\bibitem[Zheng and Wu(2024)]{uvmnet}
Zhuoran Zheng and Chen Wu.
\newblock U-shaped vision mamba for single image dehazing.
\newblock \emph{arXiv preprint arXiv:2402.04139}, 2024.

\bibitem[Zhou et~al.(2020)Zhou, Zhang, Zuo, and Loy]{IGNN}
Shangchen Zhou, Jiawei Zhang, Wangmeng Zuo, and Chen~Change Loy.
\newblock Cross-scale internal graph neural network for image super-resolution.
\newblock In \emph{Neural Information Processing Systems}, 2020.
\newblock NeurIPS 2020.

\bibitem[Zhou et~al.(2023)Zhou, Li, Guo, Bai, Cheng, and Hou]{SRFormer}
Yupeng Zhou, Zhen Li, Chun-Le Guo, Song Bai, Ming-Ming Cheng, and Qibin Hou.
\newblock Srformer: Permuted self-attention for single image super-resolution.
\newblock In \emph{IEEE Conference on Computer Vision and Pattern Recognition}, pages 12780--12791, 2023.

\bibitem[Zhu et~al.(2024)Zhu, Liao, Zhang, Wang, Liu, and Wang]{vim}
Lianghui Zhu, Bencheng Liao, Qian Zhang, Xinlong Wang, Wenyu Liu, and Xinggang Wang.
\newblock Vision mamba: Efficient visual representation learning with bidirectional state space model.
\newblock \emph{arXiv preprint arXiv:2401.09417}, 2024.

\end{thebibliography}
}


\end{document}